\documentclass{article}

\usepackage{microtype}
\usepackage{graphicx}
\usepackage{subfigure}
\usepackage{booktabs} 

\usepackage{amsmath}
\usepackage{amssymb}
\usepackage{amsfonts}       
\usepackage{nicefrac}       
\usepackage{paralist}
\usepackage{tabularx}
\usepackage{float}

\usepackage{blindtext}
\usepackage{scrextend}
\addtokomafont{labelinglabel}{\sffamily}
\usepackage{minibox}
\usepackage{amsthm}
\usepackage{enumitem}

\DeclareMathOperator*{\argmin}{arg\,min}
\DeclareMathOperator*{\argmax}{arg\,max}
\newcommand{\defeq}{\stackrel{\text{def}}{=}}

\usepackage{hyperref}

\usepackage[accepted]{icml2018}

\usepackage{algorithmic}

\icmltitlerunning{The Bottleneck Simulator: A Model-based Deep Reinforcement Learning Approach} 

\begin{document}

\twocolumn[
\icmltitle{The Bottleneck Simulator: \\ A Model-based Deep Reinforcement Learning Approach}




\icmlsetsymbol{equal}{*}

\begin{icmlauthorlist}
\icmlauthor{Iulian Vlad Serban}{to}
\icmlauthor{Chinnadhurai Sankar}{to}
\icmlauthor{Michael Pieper}{to}
\icmlauthor{Joelle Pineau}{ed}
\icmlauthor{Yoshua Bengio}{to}
\end{icmlauthorlist}

\icmlaffiliation{to}{Montreal Institute for Learning Algorithms (MILA), Montreal}
\icmlaffiliation{ed}{School of Computer Science, McGill University, Montreal and Facebook Artificial Intelligence Research (FAIR), Montreal}

\icmlcorrespondingauthor{Iulian Vlad Serban}{iulian [DOT] vlad [dot] serban [AT] umontreal [DOT] com}

\icmlkeywords{Machine Learning, ICML}

\vskip 0.3in ]



\printAffiliationsAndNotice{}  

\begin{abstract}
Deep reinforcement learning has recently shown many impressive successes.
However, one major obstacle towards applying such methods to real-world problems is their lack of data-efficiency.
To this end, we propose the Bottleneck Simulator: a model-based reinforcement learning method which combines a learned, factorized transition model of the environment with rollout simulations to learn an effective policy from few examples.
The learned transition model employs an abstract, discrete (bottleneck) state, which increases sample efficiency by reducing the number of model parameters and by exploiting structural properties of the environment.
We provide a mathematical analysis of the Bottleneck Simulator in terms of fixed points of the learned policy, which reveals how performance is affected by four distinct sources of error: an error related to the abstract space structure, an error related to the transition model estimation variance, an error related to the transition model estimation bias, and an error related to the transition model class bias.
Finally, we evaluate the Bottleneck Simulator on two natural language processing tasks: 
a text adventure game and a real-world, complex dialogue response selection task.
On both tasks, the Bottleneck Simulator yields excellent performance beating competing approaches.
\end{abstract}

\newtheorem{mytheorem}{Theorem}
\newtheorem{mycorollary}{Corollary}
\newtheorem{mydef}{Definition}
\newtheorem{mylemma}{Lemma}

\section{Introduction}
Deep reinforcement learning (RL) has recently shown impressive successes across a variety of tasks~\citep{mnih2013playing,tesauro1995temporal,silver2017mastering,silver2016mastering,brown2017superhuman,watter2015embed,lillicrap2015continuous,schulman2015high,levine2016end}.
However, the silver bullet for many of these successes have been enormous amounts of training data and result in policies which do not generalize to changes or novel tasks in the environment.
Fewer successes have been achieved outside the realm of simulated environments or environments where agents can play against themselves.
Thus, one major impediment towards applying deep RL to real-world problems is a lack of data-efficiency. 

One promising solution is \textit{model-based} RL, where an internal model of the environment is learned.
By learning an internal environment model the agent may be able to exploit structural properties of the environment.
This enables the agent to reduce the amount of trial-and-error learning and to better generalize across states and actions.


In this paper we propose a model-based RL method based on learning an approximate, factorized transition model.
The approximate transition model involves discrete, abstract states acting as information bottlenecks, which mediate the transitions between successive full states.
Once learned, the approximate transition model is then applied to learn the agent's policy (for example, using Q-learning with rollout simulations).
This method has several advantages.
First, the factorized model has significantly fewer parameters compared to a non-factorized transition model, making it highly sample efficient.
Second, by learning the abstract state representation with the specific goal of obtaining an optimal policy (as opposed to maximizing the transition model's predictive accuracy), it may be possible to trade-off some of the transition model's predictive power for an improvement in the policy's performance.
By grouping similar states together into the same discrete, abstract state, it may be possible to improve the performance of the policy learned with the approximate transition model.


The idea of grouping similar states together has been proposed before in a variety of forms (e.g.\@ state aggregation~\citep{bean1987aggregation,bertsekas1989adaptive,dietterich2000hierarchical,jong2005state,jiang2015abstraction}).
In contrast to many previous approaches, in our method the grouping is applied exclusively within the approximate transition model, while the agent's policy still operates on the complete world state.
This allows the agent's policy (e.g.\@ a neural network) to form its own high-level, distributed representations of the world from the complete world state. 
Importantly, in this method, the agent's policy is capable of obtaining better performance compared to standard state aggregation, because it may counter deficiencies in the abstract state representation by optimizing for myopic (next-step) rewards which it can do efficiently by accessing the complete world state.
This is particularly advantageous when it is possible to pretrain the policy to imitate a myopic human policy (e.g.\@ by imitating single actions or preferences given by humans) or with a policy learned on a similar task.
Furthermore, as with state aggregation, the grouping may incorporate prior structural knowledge.
As shown by the experiments, by leveraging simple knowledge of the problem domain significant performance improvements are obtained.

Our contributions are three-fold.
First, we propose a model-based RL method, called the Bottleneck Simulator, which learns an approximate transition distribution with discrete abstract states acting as information bottlenecks.
We formally define the Bottleneck Simulator and its corresponding Markov decision process (MDP) and describe the training algorithm in details.
Second, we provide a mathematical analysis based on fixed points.
We provide two upper bounds on the error incurred when learning a policy with an approximate transition distribution: one for general approximate transition distributions and one for the Bottleneck Simulator.
In particular, the second bound illustrates how the overall error may be attributed to four distinct sources: an error related to the abstract space structure (structural discrepancy), an error related to the transition model estimation variance, an error related to the transition model estimation bias, and an error related to the transition model class bias.
Finally, we demonstrate the data-efficiency of the Bottleneck Simulator on two tasks involving few data examples: a text adventure game and a real-world, complex dialogue response selection task.
We demonstrate how efficient abstractions may be constructed and show that the Bottleneck Simulator beats competing methods on both tasks.
Finally, we investigate the learned policies qualitatively and, for the text adventure game, measure how performance changes as a function of the learned abstraction structure.


\section{Background}

\subsection{Definitions}
A Markov decision process (MDP) is a tuple $\langle S, A, P, R, \gamma \rangle$, where $S$ is the set of states, $A$ is the set of actions, $P$ is the state transition probability function, $R(s, a) \in [0, r_{\text{max}}]$ is the reward function, with $r_{\text{max}} > 0$, and $\gamma \in (0, 1)$ is the discount factor~\citep{sutton1998reinforcement}.
We adopt the standard MDP formulation with finite horizon. At time $t$, the agent is in a state $s_t \in S$, takes an action $a_t \in A$, receives a reward $r_t = R(s_t, a_t)$ and transitions to a new state $s_{t+1} \in S$ with probability $P(s_{t+1} | s_t, a_t)$.



We assume the agent follows a stochastic policy $\pi$. Given a state $s \in S$, the policy $\pi$ assigns a probability to each possible action $a \in A$:
$\pi(a | s) \in [0, 1], \quad \text{s.\@ t.\@} \ \sum_{a \in A} \pi(a | s) = 1$.
The agent's goal is to learn a policy maximizing the discounted sum of rewards:
$R = \sum_{t=1}^T \gamma^t r_t$,
called the \textit{cumulative return}. or, more briefly, the \textit{return}.

Given a policy $\pi$, the \textit{state-value function} $V^{\pi}$ is defined as the expected return of the policy starting in state $s \in S$:
\begin{align}
V^{\pi}(s) = \text{E}_{\pi} \left [ \sum_{t=1}^T \gamma^t r_t \ | \ s_1 = s \right ].
\end{align}
The \textit{state-action-value function} $Q^{\pi}$ is the expected return of taking action $a$ in state $s$, and then following policy $\pi$:
\begin{align}
Q^{\pi}(s, a) = \text{E}_{\pi} \left [ \sum_{t=1}^T \gamma^t r_t \ | \ s_1 = s, a_1 = a \right ].
\end{align}
An \textit{optimal policy} $\pi^*$ is a policy satisfying $\forall s \in S, a \in A$:
\begin{align}
V^{\pi^*}(s) & = V^*(s) = \max_{\pi} V^{\pi}(s). \label{eq:optimal_value_function_definition}
\end{align}

The optimal policy can be found via dynamic programming using the Bellman optimality equations~\citep{bertsekas1995neuro,sutton1998reinforcement}, $\forall s \in S, a \in A$:
\begin{align}
V^*(s) &= \max_{a \in A} Q^*(s, a), \quad \\
Q^*(s, a) &= R(s, a) + \gamma \sum_{s' \in S} P(s' | s, a) V^*(s') \nonumber
\end{align}
which hold if and only if eq.\@ \eqref{eq:optimal_value_function_definition} is satisfied.
Popular algorithms for finding an optimal policy include Q-learning, SARSA and REINFORCE~\citep{sutton1998reinforcement}.

\subsection{Model-based RL with Approximate Transition Models}
Suppose we aim to learn an efficient policy for the MDP $\langle S, A, P, R, \gamma \rangle$, but without having access to the transition distribution $P$.
However, suppose that we still have access to the set of states and actions, the discount factor $\gamma$ and the reward function for each state-action pair $R(s, a)$.
This is a plausible setting for many real-world applications, including natural language processing, health care and robotics.

Suppose that a dataset $D = \{s^i, a^i, s'^i\}_{i=1}$ with $|D|$ tuples has been collected with a policy $\pi_D$ acting in the true (ground) MDP~\citep{sutton1990integrated,moore1993prioritized,peng1993efficient}.\footnote{Since the reward function $R(s, a)$ is assumed known and deterministic, the dataset does not need to contain the rewards.}
We can use the dataset $D$ to estimate an approximate transition distribution $P_{\text{Approx}}$:
\begin{align*}
P_{\text{Approx}}(s'|s, a) \approx P(s'|s, a) \quad \forall s, s' \in S, a \in A.
\end{align*}
Given $P_{\text{Approx}}$, we can form an approximate MDP $\langle S, A, P_{\text{Approx}}, R, \gamma \rangle$ and learn a policy $\pi$ satisfying the Bellman equations in the approximate MDP, $\forall s \in S, a \in A$:
\begin{align}
V_{\text{Approx}}(s) &= \max_{a \in A} Q_{\text{Approx}}(s, a), \label{eq:approx_mdp_bellman_equation}\\
Q_{\text{Approx}}(s, a) &= R(s, a) + \gamma \sum_{s' \in S} P_{\text{Approx}}(s' | s, a) V_{\text{Approx}}(s'), \nonumber
\end{align}
in the hope that $P_{\text{Approx}}(s' | s, a) \approx P(s' | s, a)$ implies $Q_{\text{Approx}}(s, a) \approx Q(s', a')$ $\forall s \in S, a \in A$ for policy $\pi$.

The most common approach is to learn $P_{\text{Approx}}$ by counting co-occurrences in $D$~\citep{moore1993prioritized}: 
\begin{align}
P_{\text{Approx}}(s'|s, a) = \dfrac{\text{Count}(s, a, s')}{\text{Count}(s, a, \cdot)}, \label{eq:naive_cooccurence_model}
\end{align}
where $\text{Count}(s, a, s')$ is the observation count for $(s, a, s')$ and $\text{Count}(s, a, \cdot) = \sum_{s'} \text{Count}(s, a, s')$ is the observation count for $(s, a)$ followed by any other state.
Unfortunately, this approximation is not sample efficient, because its sample complexity for accurately estimating the transition probabilities may grow in the order of $O(|S|^2 |A|)$ (see appendix).

The next section presents the Bottleneck Simulator, which learns a more sample efficient model and implements an inductive bias by utilizing information bottlenecks 
%



\section{Bottleneck Simulator}

\subsection{Definition}

We are now ready to define the Bottleneck Simulator, which is given by the tuple $\langle Z, S, A, P_{\text{Abs}}, R, \gamma \rangle$, where $Z$ is a discrete set of \textit{abstract states}, $S$ is the set of (full) states, $A$ is the set of actions and $P_{\text{Abs}}$ is a set of distributions.\footnote{In the POMDP literature, $z$ often represents the observation. However, in our notation, $z$ represents the discrete, abstract state.}
Further, we in general assume that $|Z| << |S|$.

\begin{figure}[ht]
  \centering
  \includegraphics[scale=0.19]{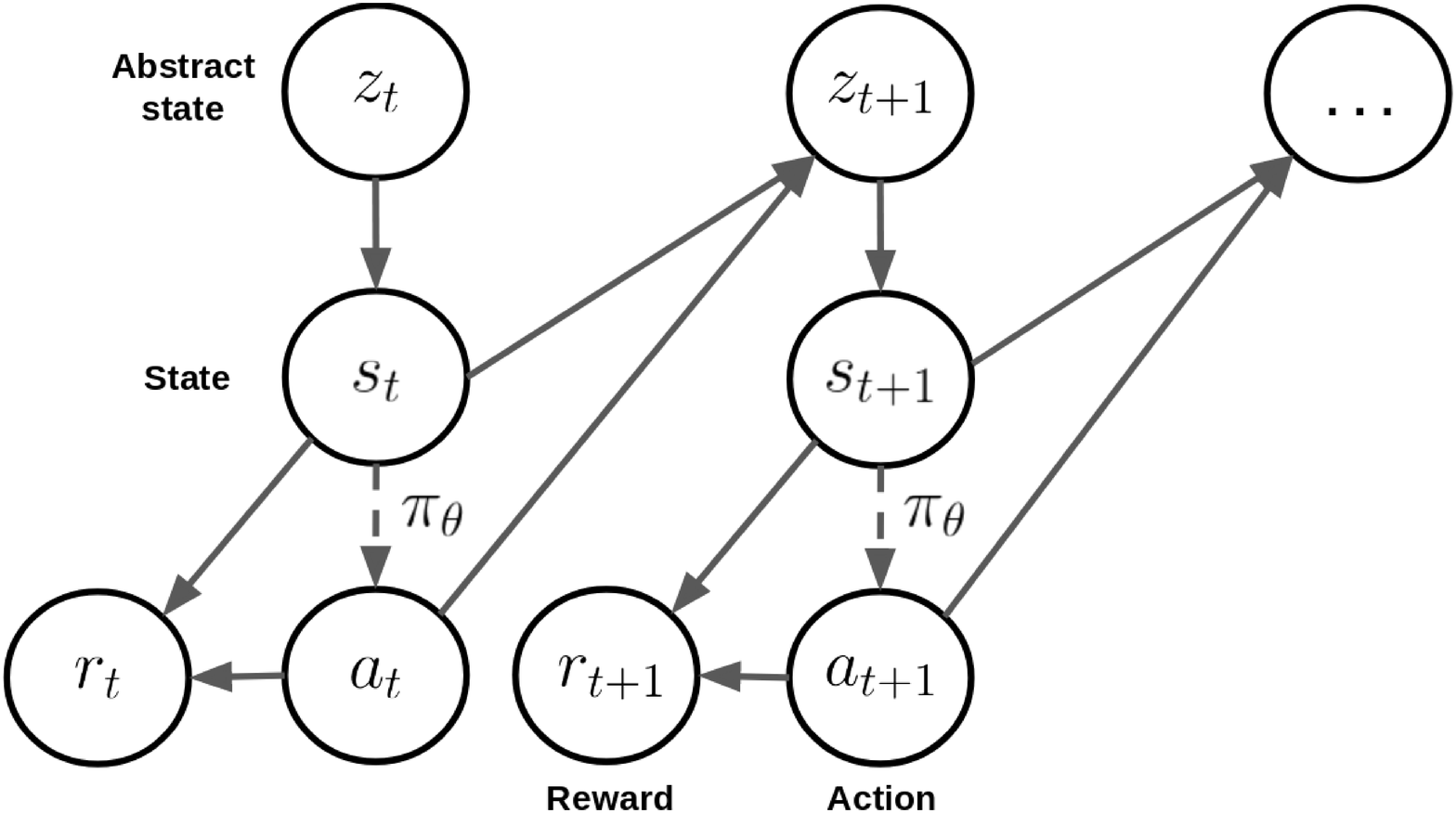}
    \caption{Probabilistic directed graphical model for the \emph{Bottleneck Simulator}. For each time step $t$, $z_t$ is a discrete random variable which represents the abstract state mediating the transitions between the successive full states $s_t$, $a_t$ represents the action taken by the agent, and $r_t$ represents the sampled reward.}
  \label{fig:approximate_mdp_second}
\end{figure}

The Bottleneck Simulator is illustrated in Figure \ref{fig:approximate_mdp_second}.
Conditioned on an abstract state $z \in Z$, a state $s \in S$ is sampled. Conditioned on a state $s$ and an action $a \in A$, a reward $r_t$ is outputted. Finally, conditioned on a state $s$ and an action $a$, the next abstract state $z' \in Z$ is sampled.
Formally, the following distributions are defined:
\begin{align}
& P_{\text{Abs}}(z_0) && \text{Initial distribution of $z$} \\
& P_{\text{Abs}}(z_{t+1} | s_t, a_t) && \text{Transition distribution of $z$} \\
& P_{\text{Abs}}(s_t | z_t) && \text{Conditional distribution of $s$} 
\end{align}

When viewed as a Markov chain, the abstract state $z$ is a \textit{Markovian} state: given a sequence $(z_1, s_1, a_1, \dots, z_{t-1}, s_{t-1}, a_{t-1}, z_t)$, all future variables depend only on $z_t$.
As such, the abstract state acts as an information bottleneck, since it has a much lower cardinality than the full states (i.e.\@ $|Z| << |S|$).
This bottleneck helps reduce sparsity and improve generalization.
Furthermore, the representation for $z$ can be learned using unsupervised learning or supervised learning on another task.
It may also incorporate domain-specific knowledge.

Further, assume that for each state $s \in S$ there exists exactly one abstract state $z \in Z$ where $P_{\text{Abs}}(s | z)$ assigns non-zero probability.
Formally, let $f_{s \to z}(s)$ be a known surjective function mapping from $S$ to $Z$, such that:
\begin{align}
P_{\text{Abs}}(s | z) = 0 \quad \forall s \in S, z \in Z \ \text{if } \ f_{s \to z}(s) \neq z.
\end{align}
This assumption allows us to construct a simple estimator for the transition distribution based on $f_{s \to z}(s)$.
Given a dataset $D = \{s^i, a^i, s'^i\}_{i=1}$ of tuples collected under a policy $\pi_D$ acting in the true MDP, we may estimate $P_{\text{Abs}}$ as:
\begin{align*}
P_{\text{Abs}}(z_0) & = 1_{(f_{s \to z}(s_{\text{start}}) = z_0)} \\
P_{\text{Abs}}(z_{t+1} | s_t, a_t) & = \dfrac{\sum_{s'; f_{s \to z}(s') = z_{t+1}} \text{Count}(s, a, s')}{\sum_{s'} \text{Count}(s, a, s')} \\
P_{\text{Abs}}(s_t | z_t) & = \dfrac{\sum_{s, a} \text{Count}(s, a, s_t)}{\sum_{s, a, s'; f_{s \to z}(s') = z_{t}} \text{Count}(s, a, s') }
\end{align*}

This approximation has a sample complexity in the order of $O(|S| |Z| |A|)$ (see appendix).
This should be compared to the estimator discussed previously, based on counting co-occurrences, which had a sample complexity of $O(|S|^2 |A|)$.
For $|Z| << |S|$, clearly $|S| |Z| |A| << |S|^2 |A|$.
As such this estimator is likely to achieve lower variance transition probability estimates for problems with large state spaces.

However, the lower variance comes at the cost of an increased bias.
By partitioning the states $s \in S$ into groups, the abstract states $z \in Z$ must contain all salient information required to estimate the true transition probabilities:
\begin{align*}
P_{\text{Abs}}(s_{t+1} | s_t, a_t) & = \sum_{z_{t+1}} P_{\text{Abs}}(z_{t+1} | s_t, a_t) P_{\text{Abs}}(s_{t+1} | z_{t+1}) \\
& \approx P(s_{t+1} | s_t, a_t).
\end{align*}
If the abstract states cannot sufficiently capture this information, then the approximation w.r.t.\@ the true transition distribution will be poor.
This in turn is likely to cause the policy learned in the Bottleneck Simulator to yield poor performance.
The same drawback applies to common state aggregation methods (e.g.\@ state aggregation~\citep{bean1987aggregation,bertsekas1989adaptive}).
However, unlike aggregation method, the policy learned with the Bottleneck Simulator has access to the complete, world state. 
Finally, it should be noted that the count-based model for $P_{\text{Abs}}$ is still rather na\"ive and inefficient.
In the next sub-section, we propose a more efficient method for learning $P_{\text{Abs}}$.

\subsection{Learning}
We assume that $f_{s \to z}$ is known.
The transition distributions can be learned using a parametric classification model (e.g.\@ a neural network with softmax output) by optimizing its parameters w.r.t.\@ log-likelihood.
Denote by $P_{\text{Abs}} = P_{\text{Abs}, \phi}$ the transition distribution of the Bottleneck Simulator parametrized by a vector of parameters $\phi$.
Formally, we aim to optimize:
\begin{align*}
& \argmax_\phi \sum_{(s_t, a_t, s_{t+1}, \cdot) \in D} \log P_{\text{Abs}}(s_{t+1} | s_t, a_t) \\
& = \argmax_\phi  \sum_{(s_t, a_t, s_{t+1}, \cdot) \in D} \log P_{\text{Abs}}(f_{s \to z}(s_{t+1}) | s_t, a_t) \\
& \quad \quad \quad \quad \quad \quad \quad \quad \quad \quad \quad  + \log P_{\text{Abs}}(s_{t+1} | f_{s \to z}(s_{t+1}))
\end{align*}
This breaks the learning problem down into two optimization problems, which are solved separately.
In the appendix we propose a method for learning $f_{s \to z}$ and $P_{\text{Abs}}$ jointly.  

\section{Mathematical Analysis}


In this section we develop two upper bounds related to the estimation error of the state-action-value function learned in an approximate MDP.
The first bound pertains to a general class of approximate MDPs and illustrates the relationship between the learned state-action-value function and the accuracy of the approximate transition distribution.
The second bound relies on the hierarchical latent structure and applies specifically to the Bottleneck Simulator.
This bound illustrates how the Bottleneck Simulator may learn a better policy by trading-off between four separate factors.


Define the \textit{true} MDP as a tuple $\langle S, A, P, R, \gamma \rangle$, where $S$ is the set of states, $A$ is the set of actions, $P$ is the \textit{true} state transition distribution, $R(h, a) \in [0, r_{\text{max}}]$ is the \textit{true} reward function and $\gamma \in (0, 1)$ is the discount factor.

Let the tuple $\langle S, A, P_{\text{Approx}}, R, \gamma \rangle$ be an approximate MDP, where $P_{\text{Approx}}$ is the transition function.
All other variables are the same as given in the \textit{true} MDP.
Let $Q_{\text{Approx}}$ satisfy eq.\@ \eqref{eq:approx_mdp_bellman_equation}.
This approximate MDP will serve as a reference for comparison.

Let the tuple $\langle Z, S, A, P_{\text{Abs}}, R, \gamma \rangle$ be the Bottleneck Simulator, where $Z$ is the discrete set of abstract states and $P_{\text{Abs}}$ is the transition function, as defined in the previous section.
All other variables are the same as given in the \textit{true} MDP.
Finally, let $Q_{\text{Abs}}$ be the optimal state-action-value function w.r.t.\@ the Bottleneck Simulator:
\begin{align*}
& Q_{\text{Abs}}(s, a) = R(s, a) + \gamma \sum_{s' \in S} P_{\text{Abs}}(s' | s, a) V_{\text{Abs}}(s') \\
& = R(s, a) + \gamma \sum_{s' \in S, z' \in Z} P_{\text{Abs}}(s' | z') P_{\text{Abs}}(z' | s, a) V_{\text{Abs}}(s')
\end{align*}
We derive bounds on the loss defined in terms of distance between $Q^*$ and sub-optimal fixed points $Q_{\text{Approx}}$ and $Q_{\text{Abs}}$:
\begin{align*}
& ||Q^*(s, a)  - Q_{\text{Approx}}(s, a)||_{\infty}, \\
& || Q^*(s, a)  - Q_{\text{Abs}}(s, a) ||_{\infty},
\end{align*}
where $||\cdot||_{\infty}$ is the infinity norm (max norm).
In other words, we bound the maximum absolute difference between the estimated return for any tuple $(s, a)$ between the approximate state-action value and the state-action value of the optimal policy. 
The same loss criteria was proposed by~\citet[Chapter 6]{bertsekas1995neuro} as well as others.

Our first theorem bounds the loss w.r.t.\@ an approximate MDP using either the total variation distance or the Kullback-Leibler divergence (KL-divergence). This theorem follows as a simple extension of existing results in the literature~\citep{ross2013interactive,talvitie2015agnostic}.
\begin{mytheorem}.
Let $Q_{\text{Approx}}$ be the optimal state-action-value function w.r.t.\@ the approximate MDP $\langle S, A, P_{\text{Approx}}, R, \gamma \rangle$, and let $Q^*$ be the optimal state-action-value function w.r.t.\@ the true MDP $\langle S, A, P, R, \gamma \rangle$.
Let $\gamma$ be their contraction rates. 
Then it holds that:
\begin{align}
& || Q^*(s, a) - Q_{\text{Approx}}(s, a) ||_{\infty} \\
& \leq  \dfrac{\gamma r_{\text{max}}}{(1 - \gamma)^2} \left | \left | \sum_{s'} \left | P(s' | s, a) - P_\text{Approx}(s' | s, a) \right | \right | \right |_{\infty} \label{eq:general_approximate_theorem} \\
& \leq  \dfrac{\gamma r_{\text{max}}\sqrt{2}}{(1 - \gamma)^2}  \left | \left | \sqrt{ D_{\text{KL}}(P(s' | s, a) || P_\text{Approx}(s' | s, a))} \right | \right |_{\infty}, \label{eq:general_approximate_theorem_kl_divergence}
\end{align}
where $D_{\text{KL}}(P(s' | s, a) || P_\text{Approx}(s' | s, a))$ is the conditional KL-divergence between $P(s' | s, a)$ and $P_\text{Approx}(s' | s, a)$.
\begin{proof}
See appendix.
\end{proof}
\end{mytheorem}

Eqs.\@ \eqref{eq:general_approximate_theorem} and \eqref{eq:general_approximate_theorem_kl_divergence} provide general bounds for any approximate transition distribution $P_\text{Approx}$ (including $P_\text{Abs}$).
The bounds are asymptotically tight in the sense that when $P_\text{Approx}$ converges to $P$ both the bounds go to zero.
Finally, the looser bound in eq.\@ \eqref{eq:general_approximate_theorem_kl_divergence} motivates why the approximate transition distribution might be learned using cross-entropy loss (or, equivalently, maximum log-likelihood).

Our second theorem bounds the loss specifically  w.r.t.\@ the Bottleneck Simulator, under the condition that if two states $s, s' \in S$ belong to the same abstract state $z \in Z$ (i.e.\@ $f_{s\to z}(s) = f_{s\to z}(s')$) then their state-value functions are close to each other w.r.t.\@ the optimal policy $\pi^*$: $|V^*(s) - V^*(s')| < \epsilon$ for some $\epsilon > 0$ if $f_{s\to z}(s) = f_{s\to z}(s')$.
This state-value similarity is closely related to metrics based on the notion of bisimulation~\citep{dean1997model,ferns2004metrics,ferns2012methods,abelhershko2016approx}.
The theorem is related to the results obtained by~\citet[p.\@ 257]{ross2013interactive} and~\citet{ross2012agnostic}, though their assumptions are different which in turn yields a bound in terms of expectations.
\begin{mytheorem}
Let $Q_{\text{Abs}}$ be the optimal state-action-value function w.r.t.\@ the Bottleneck Simulator $\langle Z, S, A, P_{\text{Abs}}, R, \gamma \rangle$, and let $Q^*$ be the optimal state-action-value function w.r.t.\@ the true MDP $\langle S, A, P, R, \gamma \rangle$. Let $\gamma$ be their contraction rates, and define:
\begin{align}
\epsilon = \max_{s_i, s_j \in S; \ f_{s\to z}(s_i) = f_{s\to z}(s_j)} |V^*(s_i) - V^*(s_j)| \label{eq:similar_abstractions_assumption}
\end{align}

Then it holds that:
\begin{align}
& ||Q^*(s, a) - Q_{\text{Abs}}(s, a)||_{\infty} \label{eq:similar_abstractions_bound_interpretable} \\
& < \dfrac{2 \gamma \epsilon}{1 - \gamma} \nonumber \\
& + \dfrac{\gamma}{1 - \gamma} \Bigg | \Bigg | \sum_{s' \in S} V_{\text{min}}(s') \Big |P_\text{Abs}(s' | s, a) -  P_\text{Abs}^{\infty}(s' | s, a) \Big | \Bigg | \Bigg |_{\infty} \nonumber \\
& + \dfrac{\gamma}{1 - \gamma} \Bigg | \Bigg | \sum_{s' \in S} V_{\text{min}}(s') \Big |P^{\infty}_\text{Abs}(s' | s, a) - P^{*}_{\text{Abs}}(s' | s, a) \Big | \Bigg | \Bigg |_{\infty} \nonumber \\
& + \dfrac{\gamma}{1 - \gamma} \Bigg | \Bigg | \sum_{s' \in S} V_{\text{min}}(s') \Big |P^{*}_\text{Abs}(s' | s, a) - P(s' | s, a) \Big | \Bigg | \Bigg |_{\infty} \nonumber
\end{align}
where $V_{\text{min}}$ and $P^{\infty}_\text{Abs}$ are defined as:
\begin{align}
V_{\text{min}}(s) &= \min_{\substack{s' \in S, \\ f_{s \to z}(s') = f_{s \to z}(s)}} V^*(s'), \\
P^{\infty}_\text{Abs}(s' | s, a) &= \sum_{z' \in Z} P^{\infty}_\text{Abs}(z' | s, a) P^{\infty}_\text{Abs}(s' | z') \\
P^{\infty}_\text{Abs}(z' | s, a) &= \sum_{s'; \ f_{s \to z}(s') = z'} P(s' | s, a)  \\
P^{\infty}_\text{Abs}(s' | z') &= \dfrac{1_{(f_{s \to z}(s') = z')} P^{\pi_D}(s')}{\sum_{\bar{s}; \ f_{s \to z}(\bar{s}) = z'} P^{\pi_D}(\bar{s})},
\end{align}
$P^{\pi_D}$ is the state visitation distribution under policy $\pi_D$, and $P^{*}_\text{Abs}$ satisfies:
\begin{align}
P^{*}_\text{Abs} = & \argmin_{\hat{P}_\text{Abs}}  \Bigg | \Bigg | \sum_{s' \in S} V_{\text{min}}(s') \Big |P(s' | s, a) -  \hat{P}_\text{Abs}(s' | s, a) \Big | \Bigg | \Bigg |_{\infty} \nonumber \\
& \ \text{s.t.\@} \ \ \hat{P}_\text{Abs}(s' | s, a) = \sum_{\substack{z' \in Z \\  f_{s \to z}(s') = z'}} \hat{P}_\text{Abs}(z' | s, a) \hat{P}_\text{Abs}(s' | z').
\end{align}

\begin{proof}
See appendix.
\end{proof}
\end{mytheorem}

The bound in eq.\@ \eqref{eq:similar_abstractions_bound_interpretable} consists of four error terms, each with an important interpretation:
\begin{align*}
|| & Q^*(s, a) - Q_{\text{Abs}}(s, a)||_{\infty} \\
 & < \text{Structural Discrepancy} \\
 & \ + \text{Transition Model Estimation Variance} \\
 & \ + \text{Transition Model Estimation Bias} \\
 & \ + \text{Transition Model Class Bias} 
\end{align*}

\textbf{Structural Discrepancy}: The \textit{structural discrepancy} is defined in eq.\@ \eqref{eq:similar_abstractions_assumption} and measures the discrepancy (or dissimilarity) between state values within each partition. 
By assigning states with similar expected returns to the same abstract partitions, the discrepancy is decreased.
Further, by increasing the number of abstract states $|Z|$ (for example, by breaking large partitions into smaller ones), the discrepancy is also decreased.
The discrepancy depends only on $Z$ and $f_{s \to z}$, which makes it independent of any collected dataset $D$.
Unlike the previous bound in eq.\@ \eqref{eq:general_approximate_theorem}, the discrepancy remains constant for $Z$ and $f_{s \to z}$.
However, in practice, as more data is accumulated it is of course desirable to enlarge $Z$ with new states.
In particular, if $|Z|$ is grown large enough such that each state belongs to its own abstract state (e.g.\@ $|Z| = |S|$) then it can be shown that this term equals zero. 

\textbf{Transition Model Estimation Variance}: This error term is a variant of the total variation distance between $P_\text{Abs}(s' |s, a)$ and $P^{\infty}_\text{Abs}(s' |s, a)$, where each term is weighted by the minimum state-value function within each abstract state $V_\text{min}(s')$.
The distribution $P^{\infty}_\text{Abs}(s' |s, a)$ represents the most accurate $P_\text{Abs}(s' | s, a)$ learned under the policy $\pi_D$ under the constraint that the model factorizes as $P^{\infty}_\text{Abs}(s' |s, a) = \sum_z P^{\infty}_\text{Abs}(s' | z') P^{\infty}_\text{Abs}(z' |s, a)$.
In other words, $P^{\infty}_\text{Abs}(s' |s, a)$ corresponds to $P_\text{Abs}(s' |s, a)$ estimated on an infinite dataset $D$ collected under the policy $\pi_D$.
As such, this error term is analogous to the variance term in the bias-variance decomposition for supervised learning problems.
Furthermore, suppose that $P_\text{Abs}^\infty = P_\text{Abs}^* = P$.
In this case, the last two error terms in the bound are exactly zero, and this error term is smaller than the general bound in eq.\@ \eqref{eq:general_approximate_theorem} since applying $V_\text{min}(s') \leq r_\text{max} / (1 - \gamma)$ yields:
\begin{align*}
& \dfrac{\gamma}{1 - \gamma} \Bigg | \Bigg | \sum_{s' \in S} V_{\text{min}}(s') \Big |P_\text{Abs}(s' | s, a) -  P_\text{Abs}^{\infty}(s' | s, a) \Big | \Bigg | \Bigg |_{\infty} \\
& \leq \dfrac{\gamma}{1 - \gamma} \Bigg | \Bigg | \sum_{s' \in S} \dfrac{r_{\text{max}}}{1 - \gamma} \Big |P_\text{Abs}(s' | s, a) -  P_\text{Abs}^{\infty}(s' | s, a) \Big | \Bigg | \Bigg |_{\infty} \\
& = \dfrac{\gamma r_{\text{max}} }{(1 - \gamma)^2}  \left | \left | \sum_{s' \in S} \left | P_\text{Abs}(s' | s, a) - P(s' | s, a) \right | \right | \right |_{\infty}
\end{align*}
For problems with large state spaces or with extreme state values, we might expect $V_\text{min}(s') << r_\text{max} / (1 - \gamma)$ for the majority of states $s' \in S$.
In this case, this bound would be far smaller than the general bound given in eq.\@ \eqref{eq:general_approximate_theorem}.
Finally, we may observe the sampling complexity of this error term.
Under the simple counting distribution introduced earlier, $P_\text{Abs}$ has $|S| |A| |Z| + |Z| |S|$ parameters.
This suggests only $O(|S| |Z| |A|)$ samples are required to reach a certain accuracy.
In contrast, for the general $P_\text{Approx}$ with a counting distribution, the sampling complexity grows with $O(|S|^2 |A|)$. 
For $|Z| << |S|$, we would expect this term to decrease on the order of $O(|S|)$ times faster than the term given in eq.\@ \eqref{eq:general_approximate_theorem}.

\textbf{Transition Model Estimation Bias}:
This error term measures the weighted total variation distance between $P^{\infty}_\text{Abs}(s' |s, a)$ and $P^*_{\text{Abs}}(s' |s, a)$, where each term is weighted by $V_\text{min}(s')$.
In other words, it measures the distance between the most accurate approximate transition distribution $P_\text{Abs}(s' | s, a)$, obtainable from an infinite dataset $D$ collected under policy $\pi_D$, and the optimal factorized transition distribution $P^*_{\text{Abs}}(s' |s, a)$ (i.e.\@ the transition distribution with the minimum sum of weighted absolute differences to the true transition distribution).
As such, this error term represents the systematic bias induced by the behaviour policy $\pi_D$.

\textbf{Transition Model Class Bias}:
This error term measures the weighted total variation distance between $P^*_{\text{Abs}}(s' |s, a)$ and $P(s' |s, a)$, where each term is weighted by $V_\text{min}(s')$.
It represents the systematic bias induced by the restricted class of probability distributions, which factorize according to latent abstract states with the mapping $f_{s \to z}$.
As such, this error term is analogous to the bias term in the bias-variance decomposition for supervised learning problems.
As more data is accumulated it is possible to enlarge $Z$ with new states.
In particular, if $|Z|$ is grown large enough, such that each state belongs to its own abstract state, and if $P_\text{Abs}(s' |s, a)$ is a tabular function, then this error term will become zero. 

The bound in eq.\@ \eqref{eq:similar_abstractions_bound_interpretable} offers more than a theoretical analysis.
In the extreme case where $|S| >> |D|$, the bound inspires hope that we may yet learn an effective policy if only we can learn an abstraction with small \textit{structural discrepancy}.

\section{Experiments}
We carry out experiments on two natural language processing tasks in order to evaluate the performance of the Bottleneck Simulator and to compare it to other approaches.
Many real-world natural language processing tasks involve complex, stochastic structures, which have to be modelled accurately in order to learn an effective policy.
Here, large amounts of training data (e.g.\@ training signals for on-policy learning, or observed trajectories of human agents executing similar tasks) are often not available.
This makes these tasks particularly suitable for demonstrating the advantages of the Bottleneck Simulator related to data-efficiency, including improved performance based on few samples.


\subsection{Text Adventure Game}
The first task is the text adventure game \textit{Home World} introduced by~\citet{2015homeworld}.
The environment consists of four connected rooms: \textit{kitchen}, \textit{bedroom}, \textit{garden} and \textit{living room}. 
The game's objective involves executing a given task in a specific room, such as \textit{eating} an \textit{apple} in the \textit{kitchen} when the task objective is \textit{"You are hungry"}.
The agent receives a reward of $1.0$ once the task is completed.
Further, we adopt the more complex setting where the objectives are redundant and confusing, such as \textit{"You are not hungry but now you are sleepy."}.
The vocabulary size is 84. The environment has 192 unique states and 22 actions.

\textbf{Setup}: We use the same experimental setup and hyper-parameters as~\citet{2015homeworld} for our baseline. 
We train an state-action-value function baseline policy parametrized as a feed-forward neural network with Q-learning.
The baseline policy is trained until the average percentage of games completed reaches $15\%$. Then, we estimate the Bottleneck Simulator environment model with the episodes collected thus far ({\raise.17ex\hbox{$\scriptstyle\sim$}}1500 transitions).
On the collected episodes, we learn the mapping from states to abstract states, $f_{s \to z}$, by applying $k$-means clustering using Euclidean distance to the word embeddings computed on the objective text and current observation text.
We use Glove word embeddings~\citep{pennington2014glove}.
We train the transition model on the collected episodes.
The transition model is a two-layer MLP classifier predicting a probability for each cluster-id of the next state given a state and action.
Finally, we train a two-layer MLP predicting the reward given a state and action.
This MLP defines the reward function in the Bottleneck Simulator environment model.

\textbf{Policy}:
We initialize the Bottleneck Simulator policy from the baseline policy and continue training it by rolling out simulations in the Bottleneck Simulator environment model.
For every 150 rollouts Bottleneck Simulator environment model, we evaluate the policy in the real game by letting the agent play out 20 episodes and measure the percentage of completed games.
We stop training when the percentage of completed games stops improving.


\textbf{Benchmark Policies}:
We compare the Bottleneck Simulator policy to two benchmark policies.
The first is the baseline policy trained with Q-learning. 
The second is a policy trained with a state abstraction method, which we call State Abstraction.
The observed states $s \in S$ are mapped to abstract states $z \in Z$, where $Z$ are the same set of abstract states utilized by the Bottleneck Simulator environment model.
As with the Bottleneck Simulator environment model, the function $f_{s \to z}$ is used to map from states to abstract states.
The action space $a \in A$ was not modified.
As with the Bottleneck Simulator policy, we evaluate the State Abstraction policy every 150 episodes by letting the agent play out another 20 episodes and measure the percentage of completed games.
For the final evaluation, we select the State Abstraction policy which obtained the highest percentage of completed games.

\textbf{Evaluation}:
The results are given in Table \ref{tabel:mdp_home_world}, averaged over 10 runs for different cluster sizes.
We evaluate the policies based on the percentage of games completed.\footnote{It should be noted that the State Abstraction policy diverged on average two out of ten times in the experiment. None of the other policies appeared to have diverged.}
It is important to note that since our goal is to evaluate sample efficiency, our policies have been trained on far fewer episodes compared to~\citet{2015homeworld}.\footnote{Indeed, a tabular state-action-value function could straightforwardly be trained with Q-learning to solve this task perfectly if given enough training examples.}
Further, we have retained the baseline policy's hyper-parameters for the Bottleneck Simulator policies while reporting the results. 
We observe peak performance at $36.8\%$ for the Bottleneck Simulator policy with a cluster size of $24$, which is significantly higher than the State Abstraction policy at $26.5\%$ and the baseline policy at only $15\%$.
This shows empirically that the Bottleneck Simulator policy is the most sample efficient algorithm, since all policies have been trained on the same number of examples.
Finally, it should be noted that the State Abstraction and Bottleneck Simulator policies are complementary and could potentially be combined (e.g.\@ by training a State Abstraction policy from samples generated by the Bottleneck Simulator environment model).



\begin{table}[t]
  \caption{Average percentage of completed games for \textit{Home World} ($\pm$ 95\% confidence intervals). Q-learning baseline policy was trained once until reaching 15\% game completion on average.} \label{tabel:mdp_home_world}
  \setlength\tabcolsep{2.5pt}
  \setlength{\extrarowheight}{2pt}
  \small
  \centering
    \begin{tabular}{ccccc}
    \toprule
    & \multicolumn{4}{c}{\textbf{Number of Clusters (i.e.\@ $|Z|$)}} \\
     \textbf{Policy} & \textbf{4} & \textbf{16} & \textbf{24} & \textbf{32} \\
    \midrule
     Q-learning (Baseline) & $15.0$ & $15.0$ & $15.0$ & $15.0$ \\
     State Abstraction & $27.8 \scriptstyle{\pm 4.4}$ & $24.7 \scriptstyle{\pm 8.4}$ & $26.5 \scriptstyle{\pm 7.5}$ & $21.3
     \scriptstyle{\pm 10.7}$\\
     Bottleneck Simulator & $17.0 \scriptstyle{\pm 2.3}$ & $24.8 \scriptstyle{\pm 2.4}$ & $\mathbf{36.8 \scriptstyle{\pm 2.1}}$ & $29.8 \scriptstyle{\pm 4.7}$\\
     \bottomrule
    \end{tabular}
\end{table}

\subsection{Dialogue} \label{subsection:dialogue_experiments}
The second task is a real-world problem, where the agent must select appropriate responses in social, chit-chat conversations.
The task is the 2017 Amazon Alexa Prize Competition~\citep{ram2017conversationalai}, where a spoken dialogue system must converse coherently and engagingly with humans on popular topics (e.g.\@ entertainment, fashion, politics, sports).\footnote{See also \url{https://developer.amazon.com/alexaprize/2017-alexa-prize}.}

\textbf{Setup}: We experiment with a dialogue system consisting of an ensemble of 22 \textit{response models}.
The response models take as input a dialogue and output responses in natural language text.
In addition, the response models may also output one or several scalar values, indicating confidence levels.
The response models have each their own internal procedure for generating responses: some response models are based on information retrieval models, others on generative language models, and yet others on template-based procedures.
Taken together, these response models output a diverse set of responses.
The dialogue system is described further in \citet{serban2017theoctopus} (see also \citet{serban2017deep}).

The agent's task is to select an appropriate response from the set of responses, in order to maximize the satisfaction of the human user.
At the end of each dialogue, the user gives a score between 1 (low satisfaction) and 5 (high satisfaction).

Prior to this experiment, a few thousand dialogues were recorded between users and two other agents acting with $\epsilon$-greedy exploration.
These dialogues are used for training the Bottleneck Simulator and the benchmark policies.
In addition, about 200,000 labels were annotated at the dialogue-turn-level using crowd-sourcing: for each recorded dialogue, an annotator was shown a dialogue and several system responses (the actual response selected by the agent as well as alternative responses) and asked to score each between 1 (very poor) and 5 (excellent).

\textbf{Policy}:
The Bottleneck Simulator policy is trained using discounted Q-learning on rollout simulations from the Bottleneck Simulator environment model.
The policy is parametrized as an state-action-value function $Q(s, a)$, taking as input the dialogue history $s$ and a candidate response $a$.
Based on the dialogue history $s$ and candidate response $a$, 1458 features are computed, including word embeddings, dialogue acts, part-of-speech tags, unigram and bigram word overlap, and model-specific features.
These features are given as input to a five-layered feed-forward neural network, which then outputs the estimated state-action value.
Further details on the model architecture are given in the appendix.

\textbf{Abstraction Space}:
As defined earlier, let $Z$ be the set of abstract states used by the Bottleneck Simulator environment model.
We then define $Z$ as the Cartesian product:
\begin{align}
Z = Z_\text{Dialogue act} \times Z_\text{User sentiment} \times Z_\text{Generic user utterance}, 
\nonumber
\end{align}
where $Z_\text{Dialogue act}$, $Z_\text{User sentiment}$ and $Z_\text{Generic user utterance}$ are three discrete sets.
The first set consists of $10$ dialogue acts, representing high-level user intentions~\citep{stolcke2000dialogue}: $Z_\text{Dialogue act} = \{\text{Accept}, \text{Reject}, \text{Request}, \text{Politics}, \text{Generic Question}, \\ \text{Personal Question}, \text{Statement}, \text{Greeting}, \text{Goodbye}, \text{Other}\}$.
These dialogue acts represent the high-level intention of the user's utterance.
The second set consists of sentiments types: $Z_\text{User sentiment} = \{\text{Negative}, \text{Neutral}, \text{Positive}\}$.
The third set contains the binary variable: $Z_\text{Generic user utterance} = \{\text{True}, \text{False}\}$.
This variable is \textit{True} only when the user utterance is generic and topic-independent (i.e.\@ when the user utterance only contains stop-words).
We develop a deterministic classifier $f_{s \to z}$ mapping dialogue histories to corresponding classes in $Z_\text{Dialogue act}$, $Z_\text{User sentiment}$ and $Z_\text{Generic user utterance}$.
Although we only consider dialogue acts, sentiment and generic utterances, it is trivial to expand the abstract state with other types of information.

\textbf{Transition Model}:
The Bottleneck Simulator environment model uses a transition distribution parametrized by three independent two-layer MLP models.
All three MLP models take as input the same features as the  Bottleneck Simulator policy, as well as features related to the dialogue act, sentiment and generic property of the last user utterance.
The first MLP predicts the next dialogue act ($Z_\text{Dialogue act}$), the second MLP predicts the next sentiment type ($Z_\text{User sentiment}$) and the third MLP predicts whether the next user utterance is generic ($Z_\text{Generic user utterance}$).
The training dataset consists of {\raise.17ex\hbox{$\scriptstyle\sim$}}$500,000$ recorded dialogue transitions, of which $70\%$ of the dialogues are used as training set and $30\%$ of the dialogues are used as validation set.
The MLPs are trained with cross-entropy using mini-batch stochastic gradient descent.
During rollout simulations, given a dialogue history $s_t$ and an action $a_t$ selected by the policy, the next abstract state $z_{t+1} \in Z$ is sampled according to the predicted probability distributions of the three MLP models.
Then, a corresponding next dialogue history $s_{t+1}$ is sampled at uniformly random from the set of recorded dialogues, under the constraint that the dialogue history matches the abstract state (i.e.\@ $f_{s\to z}(s_{t+1}) = z_{t+1}$).


\textbf{Reward Model}:
The Bottleneck Simulator environment model uses a reward model parametrized as a feed-forward neural network with a softmax output layer.
The reward model is trained to estimate the reward for each action based on the ~200,000 crowd-sourced labels. 
When rolling out simulations with the Bottleneck Simulator, the expected reward is given to the agent at each time step.
Unless otherwise stated, in the remainder of this section, this is the model we will refer to as the learned, approximate reward model.

\textbf{Benchmark Policies}:
We compare the Bottleneck Simulator policy to seven competing methods:
\begin{description}[align=left,itemindent=0.05cm,itemsep=0.05cm,font=\normalfont\emph] %
\item [Heuristic:]a heuristic policy based on pre-defined rules.
\item [Supervised:]an state-action-value function policy trained with supervised learning (cross-entropy) to predict the annotated scores on the {\raise.17ex\hbox{$\scriptstyle\sim$}}200,000 crowd-sourced labels.
\item [Q-learning:]an state-action-value function policy trained with discounted Q-learning on the recorded dialogues, where episode returns are given by a learned, approximate reward model.
\item [Q-Function Approx:]an state-action-value function policy trained on the. {\raise.17ex\hbox{$\scriptstyle\sim$}}500,000 recorded transitions with a least-squares regression loss, where the target values are given by a learned, approximate reward model.
\item [REINFORCE:]an off-policy REINFORCE policy trained with reward shaping on the {\raise.17ex\hbox{$\scriptstyle\sim$}}500,000 recorded transitions, where episode returns are given by the final user scores.
\item [REINFORCE Critic:]an off-policy REINFORCE policy trained with reward shaping on the {\raise.17ex\hbox{$\scriptstyle\sim$}}500,000 recorded transitions, where episode returns are given by a learned, approximate reward model.
\item [State Abstraction:] a tabular state-action-value function policy trained with discounted Q-learning on rollouts from the Bottleneck Simulator environment model, with abstract policy state space $Z = Z_\text{Dialogue act} \times Z_\text{User sentiment} \times Z_\text{Generic user utterance}$ containing $60$ discrete abstract states and action space containing $52$ abstract actions, and where episode returns are given by a learned, approximate reward model.
\end{description}
The two off-policy REINFORCE policies were trained with the action probabilities of the recorded dialogues (information which none of the other policies used).

With the exception of the Heuristic and State Abstraction policies, all policies were parametrized as five-layered feed-forward neural networks.
Furthermore, the Bottleneck Simulator, the Q-learning, the Q-Function Approx.\@ and the two off-policy REINFORCE policies were all initialized from the Supervised policy.
This is analogous to pretraining the policies to imitate a myopic human policy (i.e.\@ imitating the immediate actions of humans in given states).
For these policies, the first few hidden layers were kept fixed after initialization from the Supervised policy, due to the large number of parameters.
See appendix for details.

\begin{table}[t]
  \caption{Policy evaluation w.r.t.\@ average crowdsourced scores ($\pm$ 95\% confidence intervals), and average return and reward per time step computed from 500 rollouts in the Bottleneck Simulator environment model ($\pm$ standard deviations). Star $^{*}$ indicates policy is significantly better than Heuristic policy at $95\%$ statistical significance level. Triangle $^{\blacktriangle}$ indicates policy is initialized from Supervised policy feed-forward neural network and hence yield same performance w.r.t.\@ crowdsourced human scores.} \label{tabel:policy_amt_and_bottleneck_simulator_evaluation}
  \small
  \setlength\tabcolsep{2.0pt}
  \setlength{\extrarowheight}{2pt}
  \centering
    \begin{tabular}{lccc}
     \toprule
     & \textbf{Crowdsourced} & \multicolumn{2}{c}{\textbf{Simulated Rollouts}} \\ 
    \textbf{Policy} & \textbf{Human Score} & \textbf{Return} & \textbf{Avg Reward} \\
    \midrule
    \emph{Heuristic} & $2.25  \scriptstyle{\pm 0.04}$ & $-11.33  \scriptstyle{\pm 12.43}$ & $-0.29 \scriptstyle{\pm 0.19}$ \\
    \emph{Supervised} & $\mathbf{2.63  \scriptstyle{\pm 0.05}}^{*}$ & $\mathbf{-6.46 \scriptstyle{\pm 8.01}}$ & $\mathbf{-0.15 \scriptstyle{\pm 0.16}}$ \\
    \emph{Q-learning} & $\mathbf{2.63  \scriptstyle{\pm 0.05}}^{*\blacktriangle}$ & $-6.70 \scriptstyle{\pm 7.39}$ & $\mathbf{-0.15 \scriptstyle{\pm 0.17}}$ \\
    \emph{Q-Function Approx.\@} & $\mathbf{2.63  \scriptstyle{\pm 0.05}}^{*\blacktriangle}$ & $-24.19 \scriptstyle{\pm 23.30}$ & $-0.73 \scriptstyle{\pm 0.27}$ \\
    \emph{REINFORCE} & $\mathbf{2.63  \scriptstyle{\pm 0.05}}^{*\blacktriangle}$ & $-7.30 \scriptstyle{\pm 8.90}$ & $-0.16 \scriptstyle{\pm 0.16}$ \\    
    \emph{REINFORCE Critic} & $\mathbf{2.63  \scriptstyle{\pm 0.05}}^{*\blacktriangle}$ & $-10.19 \scriptstyle{\pm 11.15}$ & $-0.28 \scriptstyle{\pm 0.19}$ \\
    \emph{State Abstraction} & $1.85  \scriptstyle{\pm 0.04}$ & $-13.04 \scriptstyle{\pm 13.49}$ & $-0.35 \scriptstyle{\pm 0.19}$\\
    \emph{Bottleneck Sim.\@} & $\mathbf{2.63  \scriptstyle{\pm 0.05}}^{*\blacktriangle}$ & $-6.54 \scriptstyle{\pm 8.02}$ & $\mathbf{-0.15 \scriptstyle{\pm 0.18}}$ \\ \bottomrule
    \end{tabular}
\end{table}

\begin{table}[t]
  \caption{A/B testing experiments average real-world user scores ($\pm$ 95\% confidence intervals). Star $^{*}$ indicates policy is significantly better than other policies at $95\%$ statistical significance level. Results are based on a total of {\raise.17ex\hbox{$\scriptstyle\sim$}}3000 real-world users.} \label{tabel:ab_testing_rounds}
  \small
  \setlength\tabcolsep{4.5pt}
  \setlength{\extrarowheight}{2pt}
  \centering
    \begin{tabular}{lccc}
     \toprule
     \textbf{Policy} & \textbf{Exp 1} & \textbf{Exp 2} & \textbf{Exp 3} \\
    \midrule
    \emph{Heuristic} & $2.86 \scriptstyle{\pm 0.22}$ & - & - \\
    \emph{Supervised} & $ 2.80 \scriptstyle{\pm 0.21}$ & - & - \\
    \emph{Q-Function Approx.\@} & $ 2.74 \scriptstyle{\pm 0.21}$ & - & - \\
    \emph{REINFORCE} & $2.86 \scriptstyle{\pm 0.21}$ & $\mathbf{3.06 \scriptstyle{\pm 0.12}}$ & $3.03 \scriptstyle{\pm 0.18}$ \\
    \emph{REINFORCE Critic} & $2.84 \scriptstyle{\pm 0.23}$ & - & - \\
    \emph{Bottleneck Sim.\@} & $\mathbf{3.15 \scriptstyle{\pm 0.20}}$* & $2.92 \scriptstyle{\pm 0.12}$ & $\mathbf{3.06 \scriptstyle{\pm 0.17}}$ \\ \bottomrule
    \end{tabular}
\end{table}

\begin{table}[t]
  \caption{First A/B testing experiment topical specificity and coherence by policy. The columns are average number of noun phrases per system utterance (System NPs), average number of overlapping words between the user's utterance and the system's response (This Turn), and average number of overlapping words between the user's utterance and the system's response in the next turn (Next Turn). Stop words are excluded.  $95\%$ confidence intervals are also shown.} \label{tabel:ab_testing_round_one_topical_analysis}
  \small
  \setlength\tabcolsep{3.5pt}
  \setlength{\extrarowheight}{2pt}
  \centering
    \begin{tabular}{lccc}
     \toprule
      &  & \multicolumn{2}{c}{\textbf{Word Overlap}} \\
     \textbf{Policy} & \textbf{System NPs} & \textbf{This Turn} & \textbf{Next Turn} \\ \midrule
     \emph{Heuristic} & $1.05 \scriptstyle{\pm 0.05}$ & $7.33 \scriptstyle{\pm 0.21}$ & $2.99 \scriptstyle{\pm 1.37}$ \\
     \emph{Supervised} & $1.75 \scriptstyle{\pm 0.07}$ & $10.48 \scriptstyle{\pm 0.28}$ & $10.65 \scriptstyle{\pm 0.29}$ \\
     \emph{Q-Function Approx.\@} & $1.50 \scriptstyle{\pm 0.07}$ & $8.35 \scriptstyle{\pm 0.29}$ & $8.36 \scriptstyle{\pm 0.31}$ \\ 
     \emph{REINFORCE} & $1.45 \scriptstyle{\pm 0.05}$ & $9.05 \scriptstyle{\pm 0.21}$ & $9.14 \scriptstyle{\pm 0.22}$ \\
     \emph{REINFORCE Critic} & $1.04 \scriptstyle{\pm 0.06}$ & $7.42 \scriptstyle{\pm 0.25}$ & $7.42 \scriptstyle{\pm 0.26}$ \\
     \emph{Bottleneck Sim.\@} & $\mathbf{1.98 \scriptstyle{\pm 0.08}}$ & $\mathbf{11.28 \scriptstyle{\pm 0.30}}$ & $\mathbf{11.52 \scriptstyle{\pm 0.32}}$ \\ \bottomrule    \end{tabular}
\end{table}

\textbf{Preliminary Evaluation}:
We use two methods to perform a preliminary evaluation of the learned policies.

The first method evaluates each policy using the crowdsourced human scores.
For each dialogue history, the policy must select one of the corresponding annotated responses. Afterwards, the policy receives the human-annotated score as reward.
Finally, we compute the average human-annotated score of each policy.
This evaluation serves as a useful approximation of the immediate, average reward a policy would get on the set of annotated dialogues.\footnote{The feed-forward neural network policies were all pre-trained with cross-entropy to predict the training set of the crowdsourced labels, such that their second last layer computes the probability of each human score (see appendix for details). Therefore, the output of their last hidden layer is used to select the response in the crowdsourced evaluation. Note further that the crowdsource evaluation is carried out on the held-out test set of crowdsourced labels, while the neural network parameters were trained on the training set of crowdsourced labels.}

The second method evaluates each policy by running 500 rollout simulations in the Bottleneck Simulator environment model, and computes the average return and average reward per time step.
The rollouts are carried out on the held-out validation set of dialogue transitions (i.e.\@ only states $s \in S$, which occur in the held-out validation set are sampled during rollouts).
Although the Bottleneck Simulator environment model is far from an accurate representation of the real world, it has been trained with cross-entropy (maximum log-likelihood) on {\raise.17ex\hbox{$\scriptstyle\sim$}}500,000 recorded transitions.
Therefore, the rollout simulations might serve as a useful first approximation of how a policy might perform when interacting with real-world users.
The exception to this interpretation is the Bottleneck Simulator and State Abstraction policies, which themselves utilized rollout simulations from the Bottleneck Simulator environment model during training.
Because of this, it is possible that the these two policies might be overfitting the Bottleneck Simulator environment model and, in turn, that this evaluation might be over-estimating their performance.
Therefore, we will not consider strong performance of either of these two policies here as indicating that they are superior to other policies.

The results are given in Table \ref{tabel:policy_amt_and_bottleneck_simulator_evaluation}.
On the crowdsourced evaluation, the Supervised policy and all policies initialized perform decently reaching an average human score of $2.63$.
This is to be expected, since the Supervised policy is trained only to maximize the crowdsourced human scores.
However, the Heuristic policy performs significantly worse indicating that there is much improvement to be made on top of the pre-defined rules.
Further, the State Abstraction policy performs worst out of all the policies, indicating that the abstract state-action space cannot effectively capture important aspects of the states and actions to learn a useful policy for this complex task.
On the rollout simulation evaluation, we observe that the Supervised policy, Q-learning policy, and Bottleneck Simulator policy are tied for first place.
Since the Bottleneck Simulator policy performs similarly to the other policies here, it would appear that the policy has not overfitted the Bottleneck Simulator environment model.
After these policies follow the two REINFORCE policies and the Heuristic policy.
Second last comes the State Abstraction policy, which again indicates that the state abstraction method is insufficient for this complex task.
Finally, the Q-function Approx.\@ appears to perform the worst, suggesting that the learned, approximate reward model it was trained with does not perform well.

This section provided as a preliminary evaluation of the policies. The next section will provide a large-scale, real-world user evaluation.

\textbf{Real-World Evaluation}:
We carry out a large-scale evaluation with real-world users through three A/B testing experiments conducted during the Amazon Alexa Prize Competition, between July 29th - August 21st, 2017.
In the first experiment the Heuristic, Supervised, Q-Function Approx.\@, REINFORCE, REINFORCE Critic and Bottleneck Simulator policies were evaluated.
In the next two experiments only the Bottleneck Simulator and REINFORCE policies were evaluated.
In total, {\raise.17ex\hbox{$\scriptstyle\sim$}}3000 user scores were collected.

The average user scores are given in Table \ref{tabel:ab_testing_rounds}.
We observe that the Bottleneck Simulator policy performed best in both the first and third experiments.
This shows that the Bottleneck Simulator policy has learned an effective policy, which is in agreement with the preliminary evaluation.
On the other hand, the REINFORCE policy performed best in the second experiment.
This shows that the REINFORCE policy is the most fierce contender of the Bottleneck Simulator policy.
In line with the preliminary evaluation, the REINFORCE Critic and Q-Function Approx.\@ perform worse than the REINFORCE and Bottleneck Simulator policies.
Finally, in contrast to the preliminary evaluation, the Supervised policy performs worse than all other policies evaluated, though not significantly worse than the Heuristic policy.

Next, we conduct an analysis of the policies in the first experiment w.r.t.\@ topical specificity and topical coherence.
For topical specificity, we measure the average number of noun phrases per system utterance.
A topic-specific policy will score high on this metric.
For topical coherence, we measure the word overlap between the user's utterance and the system's response, as well as word overlap between the user's utterance and the system's response at the next turn.
The more a policy remains on topic, the higher we would expect these two metrics to be.
A good policy should have both high topical specificity and high topical coherence.

As shown in Table \ref{tabel:ab_testing_round_one_topical_analysis}, the Bottleneck Simulator policy performed best on all three metrics.
This indicates that the Bottleneck Simulator has the most coherent and engaging dialogues out of all the evaluated policies.
This is in agreement with its excellent performance w.r.t.\@ real-world user scores and w.r.t.\@ the preliminary evaluation.
A further analysis of the selected responses indicates that the Bottleneck Simulator has learned a more \textit{risk tolerant} strategy. 

\section{Related Work}

\textbf{Model-based RL}: 
Model-based RL research dates back to the 90s, and includes well-known algorithms such as Dyna, R-max and $\text{E}^3$~\citep{sutton1990integrated,moore1993prioritized,peng1993efficient,kuvayev1997approximation,brafman2002r,kearns2002near,wiering1998efficient,wiering1999reinforcement}.
Model-based RL with deep learning has also been investigated, in particular for robotic control tasks~\citep{watter2015embed,lenz2015deepmpc,gu2016continuous,finn2017deep}.
Sample efficient approaches have also been proposed by taking a Bayesian approach to learning the dynamics model. For example, PILCO incorporates uncertainty by learning a distribution over models of the dynamics in conjunction with the agent's policy~\citep{deisenroth2011pilco}.
Another approach based on Bayesian optimization was proposed by~\citet{bansal2017goal}.
An approach combining dynamics models of various levels of fidelity or accuracy was proposed by~\citet{cutler2015real}.
Other related work includes~\citet{oh2015action}, ~\citet{venkatraman2016improved},~\citet{kansky2017schema} and~\citet{racaniere2017imagination}.

The idea of grouping similar states together also has a long history in the RL community.
Numerous algorithms exists for models based on state abstraction (state aggregation)~\citep{bean1987aggregation,bertsekas1989adaptive,dean1997model,dietterich2000hierarchical,jong2005state,li2006towards,jiang2015abstraction}.
The main idea of state abstraction is to group together similar states and solve the reduced MDP. 
Solving the optimization problem in the reduced MDP requires far fewer iterations or samples, which improves convergence speed and sample efficiency.
In particular, related theoretical analyses of the regret incurred by state abstraction methods are provided by \citet{van2006performance} and \citet{petrik2014raam}.
In contrast to state abstraction, in the Bottleneck Simulator the grouping is applied exclusively within the approximate transition model while the agent's policy operates on the complete, observed state.
Compared to state abstraction, the Bottleneck Simulator reduces the impact of compounding errors caused by inaccurate abstractions in the approximate transition model.
By giving the policy access to the complete, observed state, it may counter inaccurate abstractions by optimizing for myopic (next-step) rewards.
This enables pretraining the policy to mimic a myopically optimal policy (e.g.\@ single human actions), as is the case in the dialogue response selection task.
Furthermore, the Bottleneck Simulator allows a deep neural network policy to learn its own high-level, distributed representations of the state from scratch.
Finally, the Bottleneck Simulator enables a mathematical analysis of the trade-offs incurred by the learned transition model in terms of structural discrepancy and weighted variation distances, which is not possible in the case of general, approximate transition models.

In a related vein, learning a factorized MDP (for example, by factorizing the state transition model) has also been investigated extensively in the literature~\citep{boutilier1999decision,degris2006learning,strehl2007efficient,ross2008model,bellemare2013bayesian,wu2014model,NIPS2014_5445,hallak2015off}.
For example, \citet{ross2008model} develop an efficient Bayesian framework for learning factorized MDPs.
As another example, \citet{bellemare2013bayesian} propose a Bayesian framework for learning a factored environment model based on a class of recursively decomposable factorizations. 
An important line of work in this area are stochastic factorization models~\citep{barreto2015expectation,barreto2016incremental}.
Similar to non-negative matrix factorization (NMF), these models approximate the environment transition model $\mathbf{P}$ with matrices $\mathbf{D} \mathbf{K} \approx \mathbf{P}$.
Similar to other methods, these models may improve sample efficiency when the intrinsic dimensionality of the transition model is low.
However, in comparison to the Bottleneck Simulator and other methods, it is difficult to incorporate domain-specific knowledge since $\mathbf{D}$ and $\mathbf{K}$ are learned from scratch.
In contrast to the Bottleneck Simulator and other state abstraction methods, there is no constraint for each state to belong to exactly one abstract state.
Whether or not this constraint improves or deteriorates performance is task specific.
However, without imposing this constraint, it seems unlikely that one can provide a mathematical analysis of policy performance in terms of structural discrepancy.

\textbf{Dialogue Systems}:
Numerous researchers have applied RL for training goal-oriented dialogue systems~\citep{singh1999reinforcement,williams2007partially,pieraccini2009we}.
One line of research has focused on learning dialogue systems through simulations using abstract dialogue states and actions~\citep{eckert1997user,levin2000stochastic,chung2004developing,cuayahuitl2005human,georgila2006user,schatzmann2007agenda,heeman2009representing,traum2008multi,lee2012pomdp,khouzaimi2017incremental,lopez2016automatic,su2016continuously,fatemi2016policy,asri2016sequence}.
The approaches here differ based on how the simulator is created or estimated, and whether or not the simulator is also considered an agent trying to optimize its own reward.
For example, \citet{levin2000stochastic} tackle the problem of building a flight booking dialogue system.
They estimate a user simulator model by counting transition probabilities between abstract dialogue states and user actions (similar to an n-gram model), which is then used to train an RL policy.
As a more recent example, \citet{yu2016strategy} propose to learn a dialogue manager policy through model-free off-policy RL based on simulations with a rule-based system.
Researchers have also investigated learning generative neural network policies operating directing on raw text through user simulations~\citep{zhao2016towards,guo2016learning,das2017learning,lewis2017dealornodeal,liu2017iterative}. 
In parallel to our work, \citet{peng2018integrating} have proposed a related model-based reinforcement learning approach for dialogue utilizing the Dyna algorithm.
To the best of our knowledge, the Bottleneck Simulator is the first model-based RL approach with discrete, abstract states to be applied to learning a dialogue policy operating on raw text.

\section{Conclusion}

We have proposed the Bottleneck Simulator, a model-based reinforcement learning (RL) approach combining a learned, factorized environment transition model with rollout simulations to learn an effective policy from few data examples.
The learned transition model employs an abstract, discrete state (a bottleneck state), which increases sample efficiency by reducing the number of model parameters and by exploiting structural properties of the environment.
We have provided a mathematical analysis of the Bottleneck Simulator in terms of fixed points of the learned policy.
The analysis reveals how the policy's performance is affected by four distinct sources of errors related to the abstract space structure (structural discrepancy), to the transition model estimation variance, to the transition model estimation bias, and to the transition model class bias.
We have evaluated the Bottleneck Simulator on two natural language processing tasks: a text adventure game and a real-world, complex dialogue response selection task. On both tasks, the Bottleneck Simulator has shown excellent performance beating competing approaches.
In contrast to much of the previous work on abstraction in RL, our dialogue experiments are based on a complex, real-world task with a very high-dimensional state space and evaluated by real-world users.





\bibliographystyle{icml2018}
{\bibliography{example_paper}}

\appendix

\onecolumn

\section{Dynamic Programming Preliminaries}

The Bellman optimality equations can be shortened by defining the \textit{Bellman operator} (sometimes called the \textit{dynamic programming operator}) $B$~\citep[Chapter 2]{bertsekas1995neuro}.
For a given (not necessarily optimal) state-action-value function $Q^\pi$, the operator is:
\begin{align}
(BQ^\pi)(s, a) = R(s, a) + \gamma \sum_{s' \in S} P(s' | s, a) \max_{a \in A} Q^\pi(s', a). \label{eq:bellman_operator_definition}
\end{align}
In other words, the operator $B$ updates $Q$ towards $Q^*$ with one dynamic programming iteration.

We need the following lemma, as derived by \citet{jiang2015abstraction}.
\begin{mylemma}\label{lemma:one}
Let $Q_1$ and $Q_2$ be the fixed points for the Bellman optimality operators $B_1$, $B_2$, which both operate on $\mathbb{R}^{|S| \times |A|}$ and have contraction rate $\gamma \in [0, 1)$:
\begin{align}
||Q_1 - Q_2||_{\infty} \leq \dfrac{||B_1 Q_2 - Q_2||_{\infty}}{1 - \gamma}. \label{eq:lemma_one}
\end{align}

\begin{proof}
We prove the inequality by writing out the left-hand side, applying the triangle inequality and the Bellman residual bound~\citep[Chapter 2]{bertsekas1995neuro}:
\begin{align*}
||Q_1 - Q_2||_{\infty} &= ||Q_1 - B_1 Q_2 + B_1 Q_2 - Q_2||_{\infty} \\ 
&\leq ||Q_1 - B_1 Q_2||_{\infty} + ||B_1 Q_2 - Q_2||_{\infty} \\ 
&= ||B_1 Q_1 - B_1 Q_2||_{\infty} + ||B_1 Q_2 - Q_2||_{\infty} \\ 
&\leq \gamma ||Q_1 - Q_2||_{\infty} + ||B_1 Q_2 - Q_2||_{\infty} \\ 
\end{align*}
We move the first term on the right-hand side to the other side of the inequality and re-order the terms:
\begin{align*}
||Q_1 - Q_2||_{\infty} - \gamma ||Q_1 - Q_2||_{\infty} &\leq ||B_1 Q_2 - Q_2||_{\infty} \nonumber \\
\Leftrightarrow \\ \nonumber
(1 - \gamma)||Q_1 - Q_2||_{\infty} &\leq ||B_1 Q_2 - Q_2||_{\infty} \nonumber \\
\Leftrightarrow \\ \nonumber
||Q_1 - Q_2||_{\infty} &\leq \dfrac{||B_1 Q_2 - Q_2||_{\infty}}{1 - \gamma} \nonumber 
\end{align*}

\end{proof}
\end{mylemma}

\newpage
\section{Co-occurrence Sample Efficiency}
The common, but na\"ive transition model estimated by eq.\@ \eqref{eq:naive_cooccurence_model} is not very sample efficient.
In order to illustrate this, assume that sample efficiency for a transition $(s, a, s')$ is measured as the probability that $P_{\text{Approx}}(s'|s, a)$ is more than $\epsilon > 0$ away from the true value in absolute value.
We bound it by Chebyshev's inequality:
\begin{align*}
P( | & P_{\text{Approx}}(s' | s, a)  - P(s' | s, a)| > \epsilon)) \leq \dfrac{\sigma^2}{\text{Count}(s, a, \cdot) \epsilon^2},
\end{align*}
where $\sigma^2 = \max_{s, a, s'} P(s' | s, a) (1 - P(s' | s, a))$ (i.e.\@ an upper bound on the variance of a single observation from a binomial random variable).
The error decreases inversely linear with the factor $\text{Count}(s, a, \cdot)$.
If we assume each sample $(s, a, s') \in D$ is drawn independently at uniform random, then the expected number of samples is: $\text{E}_D [ \text{Count}(s, a, \cdot) ] = |D|/(|S| |A|)$.
Summing over all $s' \in S$, we obtain the overall sample complexity:
\begin{align*}
\sum_{s'} P( | & P_{\text{Approx}}(s' | s, a)  - P(s' | s, a)| > \epsilon) \lesssim \dfrac{\sigma^2}{\epsilon^2}\dfrac{|S|^2 |A|}{|D|},
\end{align*}
which grows in the order of $O(|S|^2 |A|)$.
Unfortunately, this implies the simple model is highly inaccurate for many real-world applications, including natural language processing and robotics applications, where the state or action spaces are very large.

\newpage
\section{Theorem 1}
In this section we provide the proof for Theorem 1.
Let $Q_{\text{Approx}}$ be the optimal state-action-value function w.r.t.\@ an approximate MDP $\langle S, A, P_{\text{Approx}}, R, \gamma \rangle$, and let $Q^*$ be the optimal state-action-value function w.r.t.\@ the true MDP $\langle S, A, P, R, \gamma \rangle$.
Let $\gamma$ be their contraction rates. 
Then, the theorem states that:
\begin{align}
|| & Q^*(s, a) - Q_{\text{Approx}}(s, a) ||_{\infty} \nonumber \\
& \leq \gamma r_{\text{max}} \left | \left | \sum_{s'} \left | P(s' | s, a) - P_\text{Approx}(s' | s, a) \right | \right | \right |_{\infty} \\
& \leq \gamma r_{\text{max}} \sqrt{2} \left | \left | \sqrt{ D_{\text{KL}}(P(s' | s, a) || P_\text{Approx}(s' | s, a))} \right | \right |_{\infty}, 
\end{align}
where $D_{\text{KL}}(P(s' | s, a) || P_\text{Approx}(s' | s, a))$ is the conditional KL-divergence between $P(s' | s, a)$ and $P_\text{Approx}(s' | s, a)$:
\begin{align}
D_{\text{KL}}( P(s' | s, a)  || P_\text{Approx}(s' | s, a)) \defeq \sum_{s'} P(s' | s, a) \log \dfrac{P(s' | s, a)}{P_\text{Approx}(s' | s, a)}
\end{align}
\begin{proof}
We start by applying Lemma \ref{lemma:one} to the loss:
\begin{align}
||Q^* - Q_{\text{Approx}}||_{\infty} \leq \dfrac{1}{1 - \gamma} & || B Q_{\text{Approx}} - Q_{\text{Approx}}||_{\infty} && \text{Apply eq.\@ \eqref{eq:lemma_one}} \nonumber \\
= \dfrac{1}{1 - \gamma} & || R(s, a) + \gamma \sum_{s'} P(s' | s, a) \max_{a'} Q_\text{Approx}(s', a') && \text{Use definition in eq.\@ \eqref{eq:bellman_operator_definition}} \nonumber \\
& \ - \ Q_{\text{Approx}}(s, a) ||_{\infty}  \nonumber \\
= \dfrac{1}{1 - \gamma} & || R(s, a) + \gamma \sum_{s'} P(s' | s, a) \max_{a'} Q_\text{Approx}(s', a') && \text{Fixed point $Q_\text{Approx} = B_{\text{Approx}} Q_\text{Approx}$} \nonumber \\
& - R(s, a) - \gamma \sum_{s'} P_\text{Approx}(s' | s, a) \max_{a'} Q_\text{Approx}(s', a')||_{\infty}  \nonumber \\
= \dfrac{\gamma}{1 - \gamma} & || \sum_{s'} P(s' | s, a) \max_{a'} Q_\text{Approx}(s', a') && \text{Cancel $R(s, a)$; move $\gamma$ out} \nonumber \\
& - \sum_{s'} P_\text{Approx}(s' | s, a) \max_{a'} Q_\text{Approx}(s', a')||_{\infty}  \nonumber \\
= \dfrac{\gamma}{1 - \gamma} & \left | \left | \sum_{s'} \left ( P(s' | s, a) - P_\text{Approx}(s' | s, a) \right) \max_{a'} Q_\text{Approx}(s', a') \right | \right |_{\infty} && \text{Merge sums} \nonumber \\
\leq \dfrac{\gamma r_{\text{max}}}{(1 - \gamma)^2} & \left | \left | \sum_{s'} \left | P(s' | s, a) - P_\text{Approx}(s' | s, a) \right | \right | \right |_{\infty} && \text{Use } Q_\text{Approx}(s', a') \leq \dfrac{r_{\text{max}}}{1 - \gamma} \ \forall s', a' \nonumber
\end{align}
Here, it should be noted that norms $|| \cdot ||_{\infty}$ are taken over all combinations of $s \in S, a \in A$.
Next, we recognize the sum as being two times the total variation distance between $P(s' | s, a)$ and $P_\text{Approx}(s' | s, a)$.
Thus, we apply Pinsker's inequality \citep[p. 132]{tsybakov2009introduction} to obtain: 

\begin{align}
||Q^* - Q_{\text{Approx}}||_{\infty} & \leq \dfrac{\gamma r_{\text{max}}}{(1 - \gamma)^2} \left | \left | \sum_{s'} \left | P(s' | s, a) - P_\text{Approx}(s' | s, a) \right | \right | \right |_{\infty} \nonumber \\
& \leq \dfrac{\gamma r_{\text{max}}}{(1 - \gamma)^2} \left | \left | 2 \sqrt{\dfrac{1}{2} D_{\text{KL}}(P(s' | s, a) || P_\text{Approx}(s' | s, a))} \right | \right |_{\infty} \nonumber \\
& = \dfrac{\gamma r_{\text{max}} \sqrt{2}}{(1 - \gamma)^2} \left | \left | \sqrt{ D_{\text{KL}}(P(s' | s, a) || P_\text{Approx}(s' | s, a))} \right | \right |_{\infty} \nonumber
\end{align}
\end{proof}

\newpage
\section{Theorem 2}

In this section we provide the proof for Theorem 2.
Let $Q_{\text{Abs}}$ be the optimal state-action-value function w.r.t.\@ the Bottleneck Simulator $\langle Z, S, A, P_{\text{Abs}}, R, \gamma \rangle$, and let $Q^*$ be the optimal state-action-value function w.r.t.\@ the true MDP $\langle S, A, P, R, \gamma \rangle$. Let $\gamma$ be their contraction rates.
Finally, define:
\begin{align}
\epsilon = \max_{s_i, s_j \in S; \ f_{s\to z}(s_i) = f_{s\to z}(s_j)} |V^*(s_i) - V^*(s_j)| \label{eq:similar_abstractions_assumption_proof}
\end{align}

Then the theorem states that:
\begin{align}
& ||Q^*(s, a) - Q_{\text{Abs}}(s, a)||_{\infty} \nonumber \\
& < \dfrac{2 \gamma \epsilon}{(1 - \gamma)^2}  \\
& + \dfrac{\gamma}{(1 - \gamma)^2} \Bigg | \Bigg | \sum_{s' \in S} V_{\text{min}}(s') \Big |P_\text{Abs}(s' | s, a) -  P_\text{Abs}^{\infty}(s' | s, a) \Big | \Bigg | \Bigg |_{\infty} \nonumber \\
& + \dfrac{\gamma}{(1 - \gamma)^2} \Bigg | \Bigg | \sum_{s' \in S} V_{\text{min}}(s') \Big |P^{\infty}_\text{Abs}(s' | s, a) -  P^*_\text{Abs}(s' | s, a) \Big | \Bigg | \Bigg |_{\infty} \\ \nonumber
& + \dfrac{\gamma}{(1 - \gamma)^2} \Bigg | \Bigg | \sum_{s' \in S} V_{\text{min}}(s') \Big |P^{*}_\text{Abs}(s' | s, a) - P(s' | s, a) \Big | \Bigg | \Bigg |_{\infty} \nonumber
\end{align}
where $V_{\text{min}}$ is defined as
\begin{align}
V_{\text{min}}(s) = \min_{\substack{s' \in S, \\ f_{s \to z}(s') = f_{s \to z}(s)}} V^*(s')
\end{align}
and $P^{\infty}_\text{Abs}$ is defined as
\begin{align}
P^{\infty}_\text{Abs}(s' | s, a) &= \sum_{z \in Z} P^{\infty}_\text{Abs}(z' | s, a) P^{\infty}_\text{Abs}(s' | z') \\
P^{\infty}_\text{Abs}(z' | s, a) &= \sum_{s'; \ f_{s \to z}(s') = z'} P(s' | s, a) \\
P^{\infty}_\text{Abs}(s' | z') &= \dfrac{1_{(f_{s \to z}(s') = z')} P^{\pi_D}(s')}{\sum_{\bar{s}; \ f_{s \to z}(\bar{s}) = z'} P^{\pi_D}(\bar{s})},
\end{align}
and $P^{*}_\text{Abs}$ satisfies:
\begin{align}
P^{*}_\text{Abs} = & \argmin_{\hat{P}_\text{Abs}}  \Bigg | \Bigg | \sum_{s' \in S} V_{\text{min}}(s') \Big |P(s' | s, a) -  \hat{P}_\text{Abs}(s' | s, a) \Big | \Bigg | \Bigg |_{\infty} \nonumber \\
& \ \text{s.t.\@} \ \ \hat{P}_\text{Abs}(s' | s, a) = \sum_{\substack{z' \in Z \\  f_{s \to z}(s') = z'}} \hat{P}_\text{Abs}(z' | s, a) \hat{P}_\text{Abs}(s' | z').
\end{align}
\begin{proof}


We start by applying Lemma \ref{lemma:one} to the loss:
\begin{align*}
& ||Q^* - Q_{\text{Abs}}||_{\infty} \\
& \leq \dfrac{1}{1 - \gamma} \left | \left |Q^* - B_{\text{Abs}}Q^* \right | \right |_{\infty} && \text{Use Lemma \ref{lemma:one}} \\
& = \dfrac{1}{1 - \gamma} \Bigg | \Bigg | Q^*(s, a) - R(s, a) - \gamma \sum_{s'} P_{\text{Abs}}(s' | s, a) \max_{a'} Q^*(s', a') \Bigg | \Bigg |_{\infty} && \text{Use eq.\@ \eqref{eq:bellman_operator_definition}} \\
& = \dfrac{1}{1 - \gamma} \Bigg | \Bigg | R(s, a) - \gamma \sum_{s'} P(s' | s, a) \max_{a'} Q^*(s', a') && \text{Use $B Q^* = Q^*$} \\
& \quad \quad \quad \quad  \ - R(s, a) - \gamma \sum_{s'} P_\text{Abs}(s' | s, a) \max_{a'} Q^*(s', a') \Bigg | \Bigg |_{\infty}  \\
& = \dfrac{\gamma}{1 - \gamma} \Bigg | \Bigg | \sum_{s'} P(s' | s, a) \max_{a'} Q^*(s', a') && \text{Reorder terms, } \\
& \quad \quad \quad \quad \quad  \ - P_\text{Abs}(s' | s, a) \max_{a'} Q^*(s', a') \Bigg | \Bigg |_{\infty} && \text{  move $\gamma$ outside norm} \\
& = \dfrac{\gamma}{1 - \gamma} \Bigg | \Bigg | \sum_{s'} P(s' | s, a) V^*(s') - P_\text{Abs}(s' | s, a) V^*(s') \Bigg | \Bigg |_{\infty} && \text{Use } \max_{a'} Q^*(s', a') = V^*(s') \\
& \leq \dfrac{\gamma}{1 - \gamma} \Bigg | \Bigg | \sum_{s'} \Big | P(s' | s, a) V^*(s') - P_\text{Abs}(s' | s, a) V^*(s') \Big | \Bigg | \Bigg |_{\infty} && \text{Use } \left | \sum_i a_i \right | \leq \sum_i |a_i| \\
& = \dfrac{\gamma}{1 - \gamma} \Bigg | \Bigg | \sum_{z'} \sum_{s'; f_{s \to z}(s') = z'} \Big | P(s' | s, a) V^*(s') - P_\text{Abs}(s' | s, a) V^*(s') \Big | \Bigg | \Bigg |_{\infty} && \text{Use that } s' \text{ belongs only to one } z' \\
& = \dfrac{\gamma}{1 - \gamma} \Bigg | \Bigg | \sum_{z'} \sum_{s'; f_{s \to z}(s') = z'} \Big | P(s' | s, a) V^*(s') - P_\text{Abs}(s' | s, a) V^*(s') && \text{Add and subtract: } \\
& \quad \quad \quad \quad \quad \quad \quad \quad \quad \quad \quad \quad + P(s' | s, a) V_{\text{min}}(s') - P(s' | s, a) V_{\text{min}}(s') && \ \  P(s' | s, a) V_{\text{min}}(s') \\
& \quad \quad \quad \quad \quad \quad \quad \quad \quad \quad \quad \quad + P_\text{Abs}(s' | s, a) V_{\text{min}}(s') - P_\text{Abs}(s' | s, a) V_{\text{min}}(s') \Big | \Bigg | \Bigg |_{\infty} && \ \ P_\text{Abs}(s' | s, a) V_{\text{min}}(s') \\
& = \dfrac{\gamma}{1 - \gamma} \Bigg | \Bigg | \sum_{z'} \sum_{s'; f_{s \to z}(s') = z'} \Big | P(s' | s, a) V^*(s') - P(s' | s, a) V_{\text{min}}(s') \Big | && \text{Apply triangle inequality thrice} \\
& \quad \quad \quad \quad \quad \quad \quad \quad \quad \quad \quad + \Big | P_\text{Abs}(s' | s, a) V^*(s') - P_\text{Abs}(s' | s, a) V_{\text{min}}(s') \Big | && \ \\
& \quad \quad \quad \quad \quad \quad \quad \quad \quad \quad \quad + \Big |P(s' | s, a) V_{\text{min}}(s') -  P_\text{Abs}(s' | s, a) V_{\text{min}}(s') \Big |  \Bigg | \Bigg |_{\infty} && \\
& = \dfrac{\gamma}{1 - \gamma} \Bigg | \Bigg | \sum_{z'} \sum_{s'; f_{s \to z}(s') = z'} P(s' | s, a) \Big | V^*(s') - V_{\text{min}}(s') \Big | && \text{Combine terms} \\
& \quad \quad \quad \quad \quad \quad \quad \quad \quad \quad \quad \quad + P_\text{Abs}(s' | s, a) \Big | V^*(s') - V_{\text{min}}(s') \Big | && \\
& \quad \quad \quad \quad \quad \quad \quad \quad \quad \quad \quad \quad + V_{\text{min}}(s') \Big |P(s' | s, a) -  P_\text{Abs}(s' | s, a) \Big |  \Bigg | \Bigg |_{\infty} &&
\end{align*}

We now apply the assumption on similarity between states $s, s' \in S$ belonging to same abstract state $z' \in Z$ (eq.\@ \eqref{eq:similar_abstractions_assumption_proof}):
\begin{align*}
& \dfrac{\gamma}{1 - \gamma} \Bigg | \Bigg | \sum_{z'} \sum_{s'; f_{s \to z}(s') = z'} P(s' | s, a) \Big | V^*(s') - V_{\text{min}}(s') \Big | && \\
& \quad \quad \quad \quad \quad \quad \quad \quad \quad \quad \quad \quad + P_\text{Abs}(s' | s, a) \Big | V^*(s') - V_{\text{min}}(s') \Big | && \\
& \quad \quad \quad \quad \quad \quad \quad \quad \quad \quad \quad \quad + V_{\text{min}}(s') \Big |P(s' | s, a) -  P_\text{Abs}(s' | s, a) \Big |  \Bigg | \Bigg |_{\infty} && \\
& < \dfrac{\gamma}{1 - \gamma} \Bigg | \Bigg | \sum_{z'} \sum_{s'; f_{s \to z}(s') = z'} P(s' | s, a) \epsilon && \text{Use eq.\@ \eqref{eq:similar_abstractions_assumption} twice} \\
& \quad \quad \quad \quad \quad \quad \quad \quad \quad \quad \quad \quad + P_\text{Abs}(s' | s, a) \epsilon && \\
& \quad \quad \quad \quad \quad \quad \quad \quad \quad \quad \quad \quad + V_{\text{min}}(s') \Big |P(s' | s, a) -  P_\text{Abs}(s' | s, a) \Big |  \Bigg | \Bigg |_{\infty} && \\
& = \dfrac{\gamma}{1 - \gamma} \Bigg | \Bigg | 2 \epsilon + \sum_{z'} \sum_{s'; f_{s \to z}(s') = z'} V_{\text{min}}(s') \Big |P(s' | s, a) -  P_\text{Abs}(s' | s, a) \Big | \Bigg | \Bigg |_{\infty} && \text{Use: } \sum_{s'} P(s' | s, a) = \sum_{s'} P_\text{Abs}(s' | s, a) = 1 \\
& \leq \dfrac{2 \gamma \epsilon}{1 - \gamma} + \dfrac{\gamma}{1 - \gamma} \Bigg | \Bigg | \sum_{z'} \sum_{s'; f_{s \to z}(s') = z'} V_{\text{min}}(s') \Big |P(s' | s, a) -  P_\text{Abs}(s' | s, a) \Big | \Bigg | \Bigg |_{\infty} && \text{Use triangle inequality} \\
& = \dfrac{2 \gamma \epsilon}{1 - \gamma} + \dfrac{\gamma}{1 - \gamma} \Bigg | \Bigg | \sum_{s'} V_{\text{min}}(s') \Big |P(s' | s, a) -  P_\text{Abs}(s' | s, a) \Big | \Bigg | \Bigg |_{\infty} && \text{Contract sums} \\
& \leq \dfrac{2 \gamma \epsilon}{1 - \gamma} + \dfrac{\gamma}{1 - \gamma} \Bigg | \Bigg | \sum_{s'} V_{\text{min}}(s') \Big |P_{\text{Abs}}(s' | s, a) -  P^{\infty}_{\text{Abs}}(s' | s, a) \Big | \Bigg | \Bigg |_{\infty} && \text{Apply triangle inequality by inserting } P^{\infty}_{\text{Abs}} \\
& \quad \ \quad \ \quad \ + \dfrac{\gamma}{1 - \gamma} \Bigg | \Bigg | \sum_{s'} V_{\text{min}}(s') \Big |P(s' | s, a) -  P^{\infty}_{\text{Abs}}(s' | s, a) \Big | \Bigg | \Bigg |_{\infty} &&
\end{align*}
Finally, we apply the triangle inequality one last time by inserting $P^*_{\text{Abs}}$ in order to obtain the final result.
\end{proof}

\newpage
\section{Dialogue Experiment Benchmarks}

As discussed in the Experiments section, we compare the Bottleneck Simulator to several competing approaches. 

\subsection{Heuristic Policy} \label{subsection:heuristic}
The first approach is a heuristic policy, which selects its response from two response models in the system. The first response model is the chatbot \textit{Alice}~\citep{wallace2009anatomy,shawar2007chatbots}, which generates responses by using thousands of template rules. The second response model is the question-answering system \textit{Evi}, which is capable of handling a large variety of factual questions.\footnote{\url{www.evi.com}.}

A few pre-defined rules are used to decide if the user's utterance should be classified as a question or not. If it is classified as a question, the policy will select the response generated by the question-answering system \textit{Evi}. Otherwise, the policy will select the response generated by the chatbot \textit{Alice}.

\textit{Evi} is an industry-strength question-answering system utilizing dozens of factual databases and proprietary algorithms built over the course of an entire decade. Further, \textit{Alice} is capable of handling many different conversations effectively using its internal database containing thousands of template rules. Therefore, this policy should be considered a strong baseline.

\subsection{Supervised Policy: Learning with Crowdsourced Labels} \label{subsection:supervised_amt}
The second approach to learning a policy is based on estimating the state-action-value function using supervised learning on crowdsourced labels.
This approach also serves as initialization for the approaches discussed later.

\textbf{Crowdsourcing}: We use Amazon Mechanical Turk (AMT) to collect training data.
We follow a setup similar to~\citet{liu2016not}.
We show human evaluators a dialogue along with 4 candidate responses, and ask them to score how appropriate each candidate response is on a 1-5 Likert-type scale.
The score 1 indicates that the response is inappropriate or does not make sense, 3 indicates that the response is acceptable, and 5 indicates that the response is excellent and highly appropriate. 
The dialogues are extracted from interactions between Alexa users and preliminary versions of our system.
For each dialogue, the corresponding candidate responses are created by generating candidate responses from the 22 response models in the system.
We preprocess the dialogues and candidate responses by masking profanities and swear words. 
Furthermore, we anonymize the dialogues and candidate responses by replacing first names with randomly selected gender-neutral names.
Finally, dialogues are truncated to the last 4 utterances and last 500 words, in order to reduce the cognitive load of the task.

After the crowdsourcing, we manually inspected the annotations and observed that annotators tended to frequently overrate topic-independent, generic responses.
We corrected for this by decreasing the label scores of generic responses.

In total, we collected $199,678$ labels.
These are split into training (train), development (dev) and testing (test) datasets consisting of respectively 137,549, 23,298 and 38,831 labels each.

\begin{figure}[ht]
  \centering
  \includegraphics[scale=0.27]{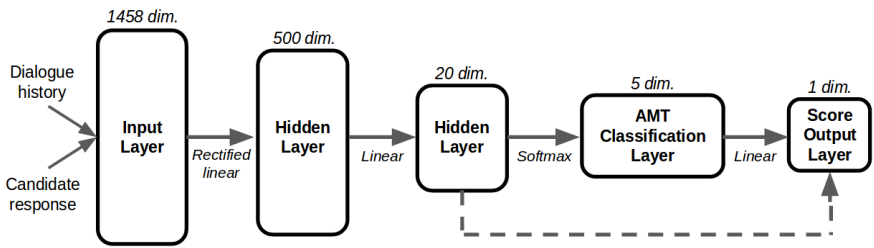}
  \caption{The system policy is parametrized as a five-layer neural network, which takes as input a dialogue history and candidate response and outputs either the estimated expected return or score. The model consists of an input layer with $1458$ features, a hidden layer with $500$ hidden units, a hidden layer with $20$ hidden units, a softmax layer with $5$ output probabilities (corresponding to the five AMT labels discussed in Section \ref{subsection:supervised_amt}), and a scalar-valued output layer. The dashed arrow indicates a skip connection.}
  \label{fig:scoring_model}
\end{figure}

\textbf{Training}: The policy is parametrized as a neural network taking as input 1458 features computed based on the dialogue history and candidate response. See Figure \ref{fig:scoring_model}.
The neural network parametrizes the state-action-value function (i.e.\@ the estimate of the expected return given a state particular state and action).
We optimize the neural network parameters w.r.t.\@ log-likelihood (cross-entropy) to predict the 4th layer, which represents the AMT label classes.
Formally, we optimize the model parameters $\theta$ as:
\begin{align}
\hat{\theta} = \argmax_{\theta} \ \sum_{x,y} \log P_\theta(y | x), \label{eq:supervised_amt_model}
\end{align}
where $x$ are the input features, $y$ is the corresponding AMT label class (a one-hot vector) and $P_\theta(y | x)$ is the model's predicted probability of $y$ given $x$, computed in the second last layer of the model.
We use the first-order gradient-descent optimizer Adam~\citep{kingma2014adampublished} .
We experiment with a variety of hyper-parameters, and select the best hyper-parameter combination based on the log-likelihood of the dev set.
For the first hidden layer, we experiment with layer sizes in the set: $\{500, 200, 50\}$.
For the second hidden layer, we experiment with layer sizes in the set: $\{50, 20, 5\}$.
We use L2 regularization on all model parameters, except for bias parameters.
We experiment with L2 regularization coefficients in the set: $\{10.0, 1.0, 10^{-1}, \dots, 10^{-9}\}$.
Unfortunately, we do not have labels to train the last layer.
Therefore, we fix the parameters of the last layer to the vector $[1.0, 2.0, 3.0, 4.0, 5.0]$.
In other words, we assign a score of $1.0$ for the label \emph{very poor}, a score of $2.0$ for the label \emph{poor}, a score of $3.0$ for the label \emph{acceptable}, a score of $4.0$ for the label \emph{good} and a score of $5.0$ for the label \emph{excellent}.
At every turn in the dialogue, the policy picks the candidate response with the highest estimated score.
As this policy was trained on crowdsourced data using supervised learning, we call it \emph{Supervised}.

\subsection{Q-learning Policy}

In the second approach, we fixed the last output layer parameters to $[1.0, 2.0, 3.0, 4.0, 5.0]$.
In other words, we assigned a score of $1.0$ for \emph{very poor} responses, $2.0$ for \emph{poor} responses, $3.0$ for \emph{acceptable} responses, and so on.
It's not clear whether this score is correlated with scores given by real-world Alexa users, which is what we ultimately want to optimize the system for.
This section describes a (deep) Q-learning approach, which directly optimizes the policy towards improving the Alexa user scores.

\textbf{Q-learning}:
Let $Q$ be the approximate, state-action-value function parametrized by parameters $\theta$.
Let $\{s_t^d, a_t^d, R^d\}_{d, t}$ be a set of observed (recorded) examples, where $s_t^d$ is the dialogue history for dialogue $d$ at time step (turn) $t$, $a_t^d$ is the agent's action for dialogue $d$ at time step (turn) $t$ and $R^d$ is the return for dialogue $d$.
Let $D$ be the number of observed dialogues and let $T^d$ be the number of turns in dialogue $d$.
Q-learning then optimizes the state-action-value function parameters by minimizing the squared error:
\begin{align}
\sum_{d=1}^{D} \sum_{t=1}^{T^d} ||Q_{\theta}(s^d_t, a^d_t) - r^d_t + \gamma \max_{a} Q_{\theta}(s^d_{t+1}, a)||^2 \label{eq:q_learning_loss}
\end{align}

\textbf{Training}:
We initialize the policy with the parameters of the \textit{Supervised} policy, and then train the parameters w.r.t.\@ eq.\@ \eqref{eq:q_learning_loss} with stochastic gradient descent using Adam.
We use a set of a few thousand dialogues recorded between users and a preliminary version of the system.
The same set of recorded dialogues were used by the Bottleneck Simulator policy.
About $80\%$ of these examples are used for training and about $20\%$ are used for development.
To reduce the risk of overfitting, we only train the parameters of the second last layer.
We select the hyper-parameters with the highest expected return on the development set.
We call this policy \emph{Q-learning}.




\subsection{Q-Function Policy}
This section describes an alternative approach to learn the state-action-value function, based on training an approximate, reward model capable of predicting the Alexa user score.

\textbf{Approximate State-Action-Value Function}:
For time (turn) $t$, let $s_t$ be a dialogue history and let $a_t$ be the corresponding response given by the system.
We aim to learn a regression model, $g_{\phi}$, which predicts the final return (user score) at the current turn:
\begin{align}
g_{\phi}(s_t, a_t) \in [1, 5],
\end{align}
where $\phi$ are the model parameters.
We call this an \textit{approximate state-action-value function} or \textit{reward model}, since it directly models the user score, which we aim to maximize.
Let $\{s_t^d, a_t^d, R^d\}_{d, t}$ be a set of observed (recorded) examples, where $t$ denotes the time step (turn) and $d$ denotes the dialogue.
Let $R^d \in [1, 5]$ denote the observed real-valued return for dialogue $d$.
The majority of users give whole number (integer) scores, but some users give decimal scores (e.g.\@ $3.5$).
Therefore, we treat $R^d$ as a real-valued number in the range $1$-$5$.
We learn the model parameters $\phi$ by minimizing the squared error between the model's prediction and the observed return:
\begin{align}
\hat{\phi} = \argmax_{\phi} \ \sum_{d} \sum_{t} (g_{\phi}(s_t^d, a_t^d) - R^d)^2
\end{align}
As before, we optimize the model parameters using mini-batch stochastic gradient descent with the optimizer Adam.
We use L2 regularization with coefficients in the set $\{10.0, 1.0, 0.1, 0.01, 0.001, 0.0001, 0.00001, 0.0\}$. We select the coefficient with the smallest squared error on a hold-out dataset.

As input to the reward model we compute 23 higher-level features based on the dialogue history and a candidate response.
In total, our dataset for training the reward model has 4340 dialogues.
We split this into a training set with 3255 examples and a test set with 1085 examples.

To increase sample efficiency, we learn an ensemble model through a variant of the bagging technique~\citep{breiman1996bagging}.
We create 5 new training sets, which are shuffled versions of the original training set.
Each shuffled dataset is split into a sub-training set and sub-hold-out set.
The sub-hold-out sets are created such that the examples in one set do not overlap with other sub-hold-out sets.
A reward model is trained on each sub-training set, with its hyper-parameters selected on the sub-hold-out set.
This increases sample efficiency by allowing us to re-use the sub-hold-out sets for training, which would otherwise not have been used.
The final reward model is an ensemble, where the output is an average of the underlying linear regression models.

\textbf{Training}:
As with the supervised learning approach, the policy here is a neural network which parametrizes an state-action-value function.
To prevent overfitting, we do not train the neural network from scratch with the reward model as target.
Instead, we initialize the model with the parameters of the \emph{Supervised} neural network, and then fine-tune it with the reward model outputs to minimize the squared error:
\begin{align}
\hat{\theta} = \argmax_{\theta} \sum_{d} \sum_{t} (f_{\theta}(s_t^d, a_t^d) - g_{\phi}(s_t^d, a_t^d))^2,
\end{align}
As before, we optimize the model parameters using mini-batch stochastic gradient descent with Adam.
As training this model does not depend on AMT labels, training is carried out on recorded dialogues.
We train on several thousand recorded dialogue examples, where about $80\%$ are used for training and about $20\%$ are used as hold-out set.
This is the same set of dialogues as were used by the Bottleneck Simulator policy.
No regularization is used.
We early stop on the squared error of the hold-out dataset w.r.t.\@ Alexa user scores predicted by the reward model.
At every turn in the dialogue, the corresponding policy picks the candidate response with the highest estimated score.
As this policy was trained with an approximate state-action-value function, we call it \emph{Q-Function Approx.\@}

We expect this policy to perform better compared to directly selecting the actions with the highest score under the reward model, because the learned policy is based on a deep neural network initialized using the crowdsourced labels.

\subsection{Off-policy REINFORCE Policy}

The previous benchmark policies parametrized the estimated state-action-value function.
Another way to parametrize the policy is as a discrete probability distribution over actions (candidate responses).
In this case, the neural network outputs real-valued scores for each candidate response. These scores are then normalized through a softmax function, such that each candidate response is assigned a probability.
This parametrization allows us to learn the policy directly from recorded dialogues through a set of methods known as \textit{policy gradient} methods.
This section describes one such approach.

\textbf{Off-policy Reinforcement Learning}: We use a variant of the classical \textit{REINFORCE} algorithm suitable for off-policy learning~\citep{williams1992simple,precup2000eligibility,precup2001off}.

As before, let $\{s_t^d, a_t^d, R^d\}_{d, t}$ be a set of observed (recorded) examples, where $s_t^d$ is the dialogue history for dialogue $d$ at time step (turn) $t$, $a_t^d$ is the agent's action for dialogue $d$ at time step (turn) $t$ and $R^d$ is the return for dialogue $d$.
Let $D$ be the number of observed dialogues and let $T^d$ be the number of turns in dialogue $d$.
Further, let $\theta_d$ be the parameters of the stochastic policy $\pi_{\theta_d}$ used during dialogue $d$.
The algorithm updates the policy parameters $\theta$ by:
\begin{align}
\Delta \theta \ \propto \ c_{t}^d \ \nabla_{\theta} \log \pi_{\theta}(a_t^d | s_t^d) \ R^d \quad \text{where} \ d \sim \text{Uniform}(1, D) \ \text{and} \ t \sim \text{Uniform}(1, T^d), \label{eq:offpolicy_reinforce}
\end{align}
where $c_{t}^d$ is the importance weight ratio:
\begin{align}
c_{t}^d \defeq \dfrac{\prod_{t'=1}^t \pi_{\theta}(a_{t'}^d | h_{t'}^d)}{\prod_{t'=1}^t \pi_{\theta_d}(a_{t'}^d | h_{t'}^d)}.
\end{align}
This ratio corrects for the discrepancy between the learned policy $\pi_{\theta}$ and the policy under which the data was collected $\pi_{\theta_d}$ (sometimes referred to as the behaviour policy).
It gives higher weights to examples with high probability under the learned policy and lower weights to examples with low probability under the learned reward function.


The importance ratio $c_{t}^d$ is known to exhibit very high, possibly infinite, variance~\citep{precup2001off}.
Therefore, we truncate the products in the nominator and denominator to only include the current time step $t$:
\begin{align}
c_{t,\text{trunc.\@}}^d \defeq \dfrac{\pi_{\theta}(a_{t}^d | h_{t}^d)}{\pi_{\theta_d}(a_{t}^d | h_{t}^d)}.
\end{align}
This induces bias in the learning process, but also acts as a regularizer.

\textbf{Reward Shaping}:
As mentioned before, one problem with the algorithm presented in eq.\@ \eqref{eq:offpolicy_reinforce} is that it suffers from high variance~\citep{precup2001off}.
The algorithm uses the return, observed only at the very end of an episode, to update the policy's action probabilities for all intermediate actions in the episode.
With a small number of examples, the variance in the gradient estimator is overwhelming.
This could easily lead the agent to over-estimate the utility of poor actions and, vice versa, to under-estimate the utility of good actions.
One remedy for this problem is \textit{reward shaping}, where the reward at each time step is estimated using an auxiliary function~\citep{ng1999policy}.
For our purpose, we propose a simple variant of reward shaping which takes into account the sentiment of the user.
When the user responds with a negative sentiment (e.g.\@ an angry comment), we will assume that the preceding action was highly inappropriate and assign it a reward of zero.
Given a dialogue $d$, at each time $t$ we assign reward $r_t^d$:
\begin{align}
r_t^d \defeq \begin{cases} 
      0 & \text{if user utterance at time $t+1$ has negative sentiment,}  \\
      \dfrac{R^d}{T^d} & \text{otherwise.}
       \end{cases}
\end{align}
With reward shaping and truncated importance weights, the learning update becomes:
\begin{align}
\Delta \theta \propto c_{t,\text{trunc.\@}}^d \nabla_{\theta} \log \pi_{\theta}(a_t^d | s_t^d) \ r_t^d \quad \text{where} \ d \sim \text{Uniform}(1, D), t \sim \text{Uniform}(1, T^d), \label{eq:offpolicy_reinforce_reward_shaping}
\end{align}

\textbf{Off-policy Evaluation}:
To evaluate the policy, we estimate the expected return under the policy~\citep{precup2000eligibility}:
\begin{align}
 \text{R}_{\pi_{\theta}}[R] \ \approx \ \sum_{d,t} c_{t,\text{trunc.\@}}^d \ r_t^d. \label{eq:offpolicy_reinforce_evaluation}
\end{align}


\textbf{Training}:
We initialize the policy with the parameters of the \textit{Supervised} policy, and then train the parameters w.r.t.\@ eq.\@ \eqref{eq:offpolicy_reinforce_reward_shaping} with stochastic gradient descent using Adam.
We use a set of a few thousand dialogues recorded between users and a preliminary version of the system.
The same set of recorded dialogues were used by the Bottleneck Simulator policy.
About $60\%$ of these examples are used for training, and about $20\%$ are used for development and about $20\%$ are used for testing.
To reduce the risk of overfitting, we only train the parameters of the second last layer.
We use a random grid search with different hyper-parameters, which include a temperature parameter and the learning rate. 
We select the hyper-parameters with the highest expected return on the development set.
We call this policy \emph{REINFORCE}.

\subsection{Off-policy REINFORCE with Learned Reward Function}
Similar to the \textit{Q-Function Approx.} policy, we may use the reward model for training with the off-policy REINFORCE algorithm.
This section describes how we combine the two approaches.

\textbf{Reward Shaping with Learned Reward Model}:
We use the reward model of the \textit{Q-Function Approx.} policy to compute a new estimate for the reward at each time step in each dialogue:
\begin{align}
r_t^d \defeq \begin{cases} 
      0 & \text{if user utterance at time $t+1$ has negative sentiment,}  \\
      g_{\phi}(s_t, a_t) & \text{otherwise.}
       \end{cases}
\end{align}
This is substituted into eq.\@ \eqref{eq:offpolicy_reinforce_reward_shaping} for training and into eq.\@ \eqref{eq:offpolicy_reinforce_evaluation} for evaluation.

\textbf{Training}:
As with the \emph{REINFORCE} policy, we initialize this policy with the parameters of the \textit{Supervised} policy, and then train the parameters w.r.t.\@ eq.\@ \eqref{eq:offpolicy_reinforce_reward_shaping} with mini-batch stochastic gradient descent using Adam.
We use the same set of dialogues and split as before.
We use a random grid search with different hyper-parameters,
As before, to reduce the risk of overfitting, we only train the parameters of the second last layer using this method.
We select the hyper-parameters with the highest expected return on the development set. 
In this case, the expected return is computed according to the learned reward model.

This policy uses the learned reward model, which approximates the state-action-value function.
This is analogous to the critic in an actor-critic architecture.
Therefore, we call this policy \emph{REINFORCE Critic}.

\subsection{State Abstraction Policy}
Finally, we describe an approach for learning a tabular state-action-value function based on state abstraction~\citep{bean1987aggregation,bertsekas1989adaptive}.

\textbf{State Abstraction}:
We define the abstract policy state space to be the same as the set of abstract states $Z$ used by the Bottleneck Simulator environment model described in Section \ref{subsection:dialogue_experiments}:
\begin{align}
Z = Z_\text{Dialogue act} \times Z_\text{User sentiment} \times Z_\text{Generic user utterance}.
\end{align}
This abstract state space contains a total of $60$ discrete states.
As with the Bottleneck Simulator environment model, the mapping $f_{s\to z}$ is used to map a dialogue history to its corresponding abstract state.

We define the abstract action space as the Cartesian product:
\begin{align}
\mathcal{A} = \mathcal{A}_{\textit{Response model class}} \times \mathcal{A}_{\textit{Wh-question}} \times \mathcal{A}_{\textit{Generic response}},
\end{align}
where $\mathcal{A}_{\textit{Response model class}}$ is a one-hot vector corresponding to one of the $13$ response model classes\footnote{Due to similarity between response models, some have been grouped together in the one-hot vector representation to reduce sparsity.} which generated the response, $\mathcal{A}_{\textit{Wh-question}} = \{True, False\}$ is a binary variable indicating whether or not the model response is a wh-question (e.g.\@ a \textit{what} or \textit{why} question), and $\mathcal{A}_{\textit{Generic response}} = \{True, False\}$ is a binary variable indicating whether the response is generic and topic-independent (i.e. a response which only contains stop-words).
The abstract action state space contains $52$ abstract actions.
A deterministic classifier is used to map an action (model response) to its corresponding abstract action.

In total, the tabular state-action-value function is parametrized by $60 \times 52 = 3120$ parameters.

\textbf{Training}:
We train the policy with Q-learning on rollouts from the Bottleneck Simulator environment model.
We use the same discount factor as the Bottleneck Simulator policy.
We call this policy \emph{State Abstraction}.

%



\end{document}